\documentclass[letterpaper,twocolumn,10pt]{article}
\usepackage{usenix-2020-09}

\usepackage{tikz}
\usepackage{amsmath}
\usepackage{xspace}
\usepackage{amssymb}
\usepackage{multirow}
\usepackage{caption}
\usepackage{subcaption}
\usepackage{nicefrac}
\usepackage{balance}
\usepackage{graphicx}
\usepackage[export]{adjustbox}
\usepackage{xcolor}
\definecolor{torstenbackground}{RGB}{208,206,206}
\usepackage[export]{adjustbox}
\usepackage{array}

\newcommand{\ournameNoSpace}{\mbox{ClearMark}}
\newcommand{\ourname}{\ournameNoSpace\xspace}
\newcommand{\ournameGen}{\ournameNoSpace's\xspace}
\newcommand{\paperTitle}{\ourname: Intuitive and Robust Model Watermarking via Transposed Model Training}

\newcommand{\bitnumber}{8,544\xspace}
\newcommand{\errorrate}{4.45\%\xspace}

\newcommand{\etal}{\emph{et~al.}\xspace}
\newcommand{\sect}{Sect.~}
\newcommand{\fig}{Fig.~}

\newcommand{\tab}{Tab.~}

\newcommand{\app}{App.~}

\newcommand{\appSect}{App.~}

\newcommand{\appFig}{App. Fig.~}

\newcommand{\resneteighteen}{\mbox{ResNet-18}\xspace}
\newcommand{\resnetthirtyfour}{\mbox{ResNet-34}\xspace}

\definecolor{darkgreen}{rgb}{0.0, 0.7, 0.0}

\definecolor{purple}{rgb}{0.5, 0.0, 0.5}

\newcommand{\cifar}{\mbox{CIFAR-10}\xspace}
\newcommand{\cifarhundred}{\mbox{CIFAR-100}\xspace}
\newcommand{\mnist}{MNIST\xspace}
\newcommand{\gtsrb}{GTSRB\xspace}

\newcommand{\naiveNoSpace}{na\"ive}
\newcommand{\naive}{\naiveNoSpace\xspace}

\newcommand{\cfNoSpace}{cf.}
\newcommand{\cf}{\cfNoSpace\xspace}

\newcommand{\dnnNoSpace}{DNN}
\newcommand{\dnn}{\dnnNoSpace\xspace}

\begin{document}

\date{}

\title{\Large \bf \paperTitle}
\author{
{\rm Torsten Krauß}\\
University of Würzburg
\and
{\rm Jasper Stang}\\
University of Würzburg
\and
{\rm Alexandra Dmitrienko}\\
University of Würzburg
} 

\maketitle

\begin{abstract}
Due to costly efforts during data acquisition and model training, Deep Neural Networks (DNNs) belong to the intellectual property of the model creator. Hence, unauthorized use, theft, or modification may lead to legal repercussions. Existing DNN watermarking methods for ownership proof are often non-intuitive, embed human-invisible marks, require trust in algorithmic assessment that lacks human-understandable attributes, and rely on rigid thresholds, making it susceptible to failure in cases of partial watermark erasure.

This paper introduces \ourname, the first DNN watermarking method designed for intuitive human assessment. \ourname embeds visible watermarks, enabling human decision-making without rigid value thresholds while allowing technology-assisted evaluations. \ourname defines a transposed model architecture allowing to use of the model in a backward fashion to interwove the watermark with the main task within all model parameters. Compared to existing watermarking methods, \ourname produces visual watermarks that are easy for humans to
understand without requiring complex verification algorithms or strict thresholds. The watermark is embedded within all model parameters and entangled with the main task, exhibiting superior robustness. 
It shows an \bitnumber-bit watermark capacity comparable to the strongest existing work. Crucially, \ournameGen effectiveness is model and dataset-agnostic, and resilient against adversarial model manipulations, as demonstrated in a comprehensive study performed with four datasets and seven architectures.
\end{abstract}

\section{Introduction}
\label{sec:intro}
In the realm of rapidly advancing technologies, machine learning stands out for many advantages, primarily revolving around automation, informed decision-making, and insightful recommendations drawn from historical data. This transformative technology finds extensive application in diverse real-world scenarios, ranging from critical tasks such as medical image classification~\cite{cai2020review} to facilitating processes like natural language processing~\cite{collobert2011natural}, and extends further into domains like autonomous driving~\cite{chen2015deepdrivingautonomous} and translation services~\cite{stahlberg2020neural}. At its core, a machine learning system consists of a model, often a Deep Neural Network (\dnn), and data. Acquiring this data, especially in large quantities necessary for robust machine learning, can be both challenging and costly. Furthermore, the data can be inherently sensitive, such as in the case of medical images, warranting strict protection under the umbrella of intellectual property (IP) rights belonging to its owner. Once data becomes available, significant computational resources are utilized for model training, which entails substantial effort and, consequently, costs. The resulting model, infused with knowledge distilled from underlying training algorithms and exhaustive data, represents a culmination of innovation and expertise in the field DNNs and embodies the IP of its creator. 

However, in case a model is made available to the public or sold to third parties, the matter of safeguarding IP rights becomes a challenge. In the face of potential adversarial entities, these models are vulnerable to theft, unauthorized resale, unwarranted modification, or illicit utilization. Such circumstances, recognized as copyright infringements, have the potential to escalate into legal proceedings, underscoring the significance of robust and comprehensive protective measures. 

The state-of-the-art approach to safeguarding the IP of trained models is model watermarking~\cite{wm_survey_one,wm_survey_two}. Existing DNN watermarking methods commonly require a key as input to an algorithm for watermark extraction. This process generates an output that is subsequently verified against the secret ground truth of the watermark. Watermarking methods can be categorized based on two main properties: whether they operate by accessing the model's internals (white-box) \cite{uchida2017embedding,li2021spreadtransform,rouhani2019deepsigns,tartaglione2021delving,wang2020watermarkingbackpropagation,wang2021riga,rouhani2019deepsigns}, such as the weights, or by analyzing only prediction outputs (black-box) \cite{lemerrer2019frontierstitching,adi2018turning,zhang2018protecting,zhang2020modelwatermarking,zheng2019blindwatermark,guo2018watermarking,jia2021entangledwm}. Moreover, a watermark is either 1-bit \cite{lemerrer2019frontierstitching,tartaglione2021delving,jia2021entangledwm} or multi-bit \cite{uchida2017embedding,li2021spreadtransform,rouhani2019deepsigns,adi2018turning,zhang2018protecting,zhang2020modelwatermarking,zheng2019blindwatermark,guo2018watermarking,wang2020watermarkingbackpropagation,wang2021riga,jia2021entangledwm,rouhani2019deepsigns}. A 1-bit watermark merely indicates the presence or absence of a watermark, whereas a multi-bit watermark incorporates details like the name of the copyright holder.

\vspace{0.1cm}
\noindent\textbf{Problem Statement.} However, the watermarking techniques proposed so far can be considered non-intuitive, as they embed human-invisible watermarks. Consequently, for watermark verification, the evaluator must place trust in the algorithm, which typically extracts data from the DNN that lack human-understandable attributes. This data is then distilled into a single value, and then compared to a rigid threshold, where values above it confirm watermark's existence and values below it deem the watermark invalid. In cases of partial watermark erasure algorithmic assessment might fall short, while humans with a clear understanding of its prior presence might be able to easily verify a copyright infringement. 
In the context of an image, for instance, it is imaginable that a watermark, depicted as a stamp, could be erased, for instance, up to 70\%; nevertheless, even with this level of reduction, it may still remain conspicuously apparent that the watermark was once embedded into the image. 

This paper addresses these challenges and introduces \ourname, the first multi-bit, white-box DNN watermarking method that follows an intuitive approach. \ourname embeds a visible watermark that can be assessed through human inspection, empowering humans to make decisions based on common sense and reasoning while retaining the option for technology-assisted evaluations. In addition, the method is generic and can be applied to various datasets and model architectures, and is robust against a wide range of watermark erasure techniques, including manipulation attempts by adaptive attackers who know the details of the protection method and even of underlying keys. 

\vspace{0.1cm}
\noindent\textbf{Contributions.} In particular, this paper makes the following contributions:
\begin{itemize}
    \item We propose \ourname, the first DNN watermarking mechanism that yields human-visible and understandable outputs, bypassing the need for rigid value thresholds and enabling more coherent and defensible jurisdiction, especially in cases where portions of the watermark remain obviously discernible. As a consequence, \ourname provides more security for the model creators' intellectual property.

    \item We have pioneered the utilization of transposed model training inspired by deconvolutions~\cite{zeiler2010deconvolutional,pixeltransposed,im2019dt,autoencoderone,deconvgan,deconvgantwo} as a novel approach for integrating a visible and human-comprehensible watermark into a DNN. In this method, we define a transposed model architecture specifically tailored to the existing model's structure, which shares weights with the original model. This design enables the training of the model in a reverse fashion, focusing on embedding the watermark, while the effects on conventional forward training for the model's primary task are negligible.

    \item We entangle the watermark and the main task within all model layers to create a robust watermark that can withstand adversarial model manipulations, such as fine-tuning on third-party datasets, pruning of model parameters, or adaptive adversaries attempting to remove or overwrite the watermark. Inspired by Siamese Networks~\cite{koch2015siamese}, we achieve this by sharing weights between the model's primary task and the watermark and by enforcing that the watermark does not cause abnormal model parameters. Any malicious modification necessitates sacrifices in the model's main task performance, rendering the model less useful.
    
    \item We conduct a systematic large-scale study to analyze factors influencing \ourname, demonstrating its independence from application-specific factors by leveraging different datasets (\mnist~\cite{mnist}, \cifar~\cite{cifar}, \gtsrb~\cite{gtsrb}, and \cifarhundred~\cite{cifar}) and model architectures (CNNs, \resneteighteen, \resnetthirtyfour~\cite{resnet}, ViT~\cite{vit}, and VGG11~\cite{vgg_eleven}) during evaluation. Additionally, we test the watermark's robustness under various fine-tuning and pruning scenarios and showcase a substantial watermark capacity of \bitnumber~bits that can be embedded with a low error rate of \errorrate, which is comparable to the strongest existing work in terms of capacity with 8,400 bits~\cite{li2021spreadtransform}. 
\end{itemize}
In summary, this work introduces a highly intuitive and easy-to-understand white-box multi-bit DNN watermarking method, offering a robust defense against various attack scenarios. The embedded watermark is human-understandable, addressing the limitations of existing solutions, which often lack intuitive design and human-friendliness in the final decision-making process, a task that can be seamlessly undertaken by a human evaluator when employing \ourname.

\section{Background}
\label{sec:background}
This section provides background information that is necessary to understand our approach. 

\vspace{0.1cm}
\noindent\textbf{Watermarking}
\label{sec:background:watermarking}
Watermarking~\cite{furht2004multimedia,hartung1999multimedia,qu2007intellectual,cox1997secure,lu2004multimedia,van1994digital,katzenbeisser2016information} is a technique to embed a digital mark or identifier into digital media, e.g., images or audio, without significantly altering the content's appearance or functionality, with the purpose of embedding a discernible sign of ownership, authenticity, or other relevant information. Those signs can then be used for various scenarios, e.g., copyright protection, authenticity verification, ownership attribution, digital rights management, tamper detection, or metadata embedding. Watermarks come in various forms, with visible and concealed watermarks being notable categories. 1-bit watermarks make explicit ownership claims through their mere presence, while multi-bit watermarks convey additional information. The choice of watermarking method depends on the specific use case and the desired level of security. Extracting the watermark, which is a \textit{secret value}, relies on knowing the \textit{extraction method} and is often accompanied by a specific \textit{secret key}.

In DNNs the watermark is embedded typically within the model's parameters~\cite{wm_survey_one,wm_survey_two}. This can transpire either directly, necessitating white-box access to the parameters for extraction and verification, or indirectly through a learning process that configures the parameters to produce outputs containing the watermark when subjected to specific inputs. In such a case, black-box access is necessary for watermark verification. Both methods are achieved by introducing an additional regularization term to the loss function during model training. This regularization term orchestrates parameter adjustments in alignment with the intended watermarking objectives. 

\begin{figure}[tb]
  \centering
  \includegraphics[width=\linewidth]{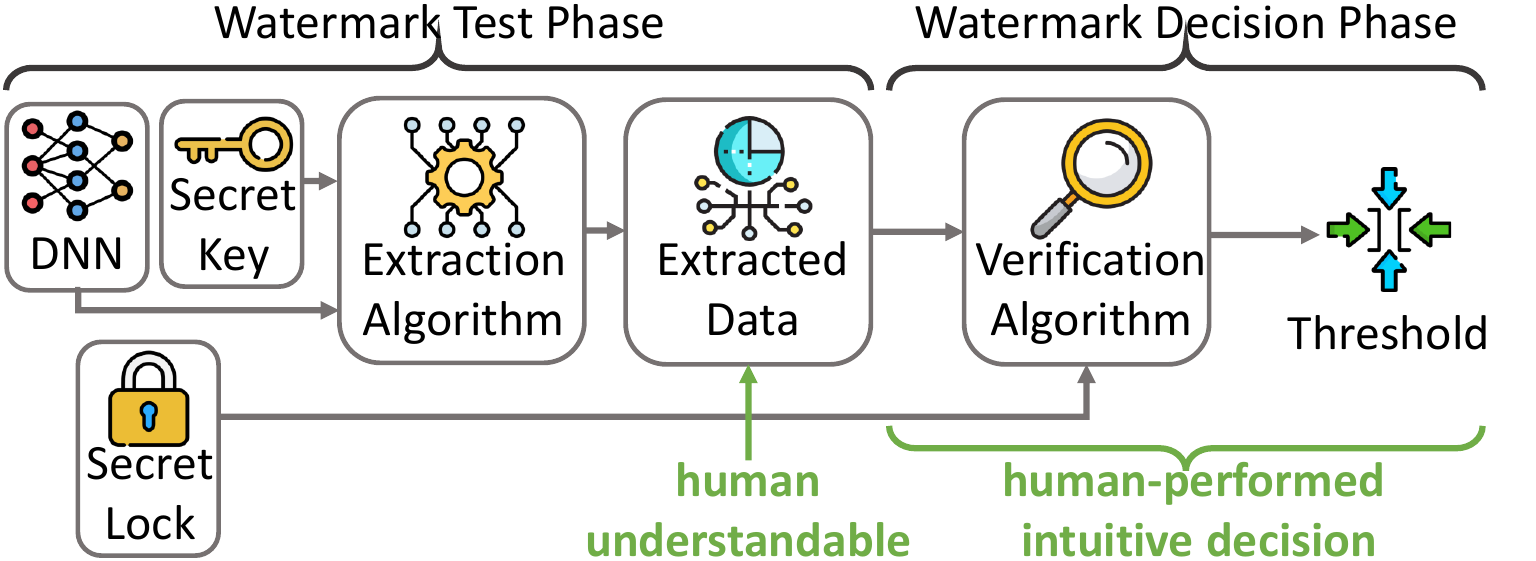}
  \caption{This Figure depicts a high-level overview of watermarking, including the test and decision phase. The green text indicates the focus of this paper.}
\label{fig:app:wmhighleveloverview}
\end{figure}

For watermark extraction and verification, as visualized in \hyperref[fig:app:wmhighleveloverview]{\fig\ref{fig:app:wmhighleveloverview}}, the \dnn and the watermark's secret key are fed into an extraction algorithm which yields some data. Those data are then verified against the watermarks ground truth secret by an algorithm that relies on a rigid threshold.

\vspace{0.1cm}
\noindent\textbf{Transposed Model Functionality}
\label{sec:background:transposed}
The transposed functionality of a machine-learning model is the reverse of the original model's task. We leverage the concept of transposed models to embed a watermark into the model. For instance, consider a model that classifies image inputs into a feature vector indicating detected objects within the image. The transposed functionality would involve inputting such a feature vector into the model to generate an image that embodies the characteristics encoded within that feature vector. This concept extends to the more granular level of model layers within the model architecture, where transposed layers reverse the functionality of their corresponding original layers. A transposed model consists of multiple transposed layers in the reverse order of the original model's architecture. Below, we discuss common model layers and existing or straightforward methods for transposing their functionality. However, certain layers may not be capable of precisely recovering the original input, especially when input data have undergone compression, resulting in information loss.

\vspace{0.1cm}
\noindent \emph{Linear Layer.} A linear layer performs a calculation such as $y = x \cdot w^T + b$, where $w$ and $b$ denote weights and bias matrices, and $x$ and $y$ represent input and output, respectively. Here, $T$ denotes the transposed operation such that the matrix is flipped over its diagonal, e.g., $A_{ij}$ becomes $A_{ji}$. Linear layers can be accurately reversed by computing $x = (y - b) \cdot w$, effectively retrieving the original input from the output.
    
\vspace{0.1cm}
\noindent \emph{Batch Normalization.} Batch normalization layers~\cite{batchnorm} are used to keep the data flowing through a model in a specific range. Such layers perform a computation akin to $y = \frac{x - E(x)}{\sqrt{Var(x) + \epsilon}} \cdot \gamma + \beta$, where $E(x)$ and $Var(x)$ are the feature-wise mean and variance of the input data, $\epsilon$ is a small constant, and $\gamma$ and $\beta$ are learnable parameters. As mean and variance are dependent on $x$, the operation cannot be reversed straightforwardly when only provided with $y$. To address this, one can set default values for $E(x) = 0$ and $Var(x) = 1$, resulting in $x = \frac{(y - \beta) \cdot \epsilon}{\gamma}$, which provides a good approximation. 

\vspace{0.1cm}
\noindent \emph{Pooling Layer.} Pooling layers~\cite{pooling} reduce the dimensionality of data by selecting representative values among multiple data points based on specific rules, such as computing the average or selecting the maximum value. As this process fundamentally involves downsampling, the transposed functionality is centered around upsampling. Consequently, the exact transposition of such downsampling computations can only be approximated, as new data points must be inferred. One common approach to upsample data is through interpolation, with several interpolation methods available, such as the nearest-neighbor algorithm \cite{Rukundo2012_nearestneighborinterpolation} or the bilinear algorithm \cite{rukundo2014bilinearinterpolation}.

\vspace{0.1cm}
\noindent \emph{Convolutional Layer.} A convolution~\cite{convolution} transforms the input to extract relevant features. This transformation applies a filter or kernel to the input to produce output data points. Multiple input data points are convolved with the filter to generate a single output data point. Hence, similar to downsampling, the computation cannot be exactly reversed. However, a method proposed in~\cite{zeiler2010deconvolutional} offers a reasonably effective approximation\footnote{The method of~\cite{zeiler2010deconvolutional} is also used in PyTorch as transposed convolution modules and can be seen as the gradient of the respective convolution.}.

\noindent \emph{Dropout Layer.} Dropout layers~\cite{dropout} are designed to distribute knowledge across various parameters. They implement a regularization technique that simulates training numerous neural networks with varied architectures concurrently. During training, random layer outputs are ignored, altering the layer's appearance and connectivity. Each training update reflects a distinct "view" of the layer. They exert influence during training, have no learnable parameters, and remain inconsequential during inference. Therefore, dropout layers are utilized identically to the forward pass during transposed training.

\noindent \emph{Activation Functions.} Activation functions, e.g., ReLU~\cite{agarap2018deep_relu}, introduce non-linearity in a model and play a significant role in the model's ability to generalize learned knowledge. Some activation functions introduce lossiness, like ReLU, which maps all negative input values to zero while the positive input values remain the same. Naturally, such operations are irreversible, and thus, activation functions can not be transposed. %The activation functions in a DNN do not contain knowledge, specifically learnable parameters, but allow generalization through the introduction of non-linearity.

\noindent \emph{Transformer Blocks.} Transformer models, like Vision Tranformer~\cite{vit}, deviate from convolution-based models and are constructed by so-called transformer blocks, which consist of an encoder and an attention module, both featuring linear layers, dropout layers, and activation functions. Vision Transformer divides the image into patches and embeddings are generated for each patch. The transposed functionality of these linear layers, dropout layers and activation functions was already elaborated in the previous paragraphs, indicating that transposing a Vision Transformer is straightforward.

\vspace{0.1cm}
\noindent In summary, the transposition of model functionalities can be applied to a variety of common ML architectures. While some layers allow for a straightforward reversal, others necessitate approximation methods due to inherent complexities and information loss.

\vspace{0.1cm}
\noindent\textbf{Image Similarity}
\label{sec:background:ssim}
Human discernment of whether two images share identical content is typically straightforward. Nevertheless, conventional machine-based methods for quantifying errors, such as the computation of Mean Squared Error (MSE) across all pixels in an image, can yield substantial error values, particularly when the images possess matching structural elements but differ in aspects like color. Human visual perception excels at extracting structural information from images, and in this context, Wang~\etal\cite{ssim} introduced the Structural Similarity Index (SSIM). Within this paper, we employ SSIM as part of a loss function to embed a watermark into a DNN, as well as for post-extraction verification of the watermark's presence and integrity. The SSIM has a value range of $[-1, 1]$, where 1 indicates perfect similarity, 0 indicates no similarity, and -1 indicates perfect anti-correlation.
 
SSIM addresses the limitations of other metrics by providing a quantifiable measure of image dissimilarity considering luminance, contrast, and structure, aligning more closely with human perception. As exemplified in \hyperref[fig:ssim:background]{\tab\ref{fig:ssim:background}}, MSE calculations often highlight substantial disparities from the original image, whereas SSIM reliably identifies high levels of structural similarity of the content. As visualized in the fifth column of \hyperref[fig:ssim:background]{\tab\ref{fig:ssim:background}}, even SSIM values of, e.g., 0.18, are sufficient, such that a human can claim similarity between two images.

\section{Problem Setting}
\label{sec:problem}
\noindent \textbf{Considered Scenario.} We consider a classical watermarking scenario: The model owner trains a Deep Neural Network (DNN) by utilizing a proprietary dataset and costly resources making it critical to protect the resulting model, which is the intellectual property of the owner. Thereby, a maximally effective watermark should be embedded that allows ownership claim. The produced model is then legally or illegally distributed, e.g., sold or stolen, and placed into production. If the owner suspects a copyright infringement of their model, an inspection of the suspected model can be conducted. Precisely, it should be possible to extract and verify the watermark, even if benign or adversarial modifications have been performed on the copy of the original model. In the following, we first define the objectives (\hyperref[sec:problem:objectives]{\sect\ref{sec:problem:objectives}}) and the threat model (\hyperref[sec:problem:threatmodel]{\sect\ref{sec:problem:threatmodel}}).

\vspace{-0.2cm}
\subsection{Watermark Objectives}
\label{sec:problem:objectives}
Inspired by related works~\cite{rouhani2019deepsigns,chen2019deepmarks,xie2021deepmark,guo2018watermarking,xiaoyu2021ipguard,chen2022deepjudge,zheng2019blindwatermark,yang2022metafinger}, we define several objectives, that should be fulfilled by an effective watermark: 1)~\textit{Understandability}: The evaluation of the watermark should be easily possible by human inspection. We add this objective to the commonly used ones, as the final decision in common DNN watermarking methods typically relies on a rigid threshold that may not detect leftovers after watermark erasure attempts, but those could be  obvious to detect for human observers. Further, decisions that can be made by humans are more intuitive and easily comprehensible than empirically determined thresholds. 2)~\textit{Fidelity}: Watermark embedding should preserve the model performance on the primary task. 3)~\textit{Reliability}: It must be reliably possible to extract the watermark from previously watermarked models\footnote{Note that watermarks from models that have been significantly manipulated to the extent that the original main task is severely compromised do not require detection, as the model no longer retains the creator's IP rights.}. 4)~\textit{Robustness}: The embedded watermark must withstand model modifications, which we describe in \hyperref[sec:problem:threatmodel]{\sect\ref{sec:problem:threatmodel}}. 5)~\textit{Integrity}: The watermark method should uniquely identify the watermark's secret value respective to the watermark's key and should not extract a valid watermark from unwatermarked models. 6)~\textit{Capacity}: The amount of information embedded into the watermark should be maximized to strengthen ownership claims. 7)~\textit{Efficiency}: Embedding the watermark should introduce negligible computational overhead. 8)~\textit{Security}: The watermark should not introduce obvious footprints allowing for easy detection and removal. 9)~\textit{Generalizability}: The approach should be independent of the dataset or model architecture.

\begin{table}[tb]
\centering
  \setlength\extrarowheight{0pt}
    \begin{adjustbox}{max width=0.45\textwidth}
\begin{tabular}{c|cccccc}
Image  & \includegraphics[scale=3,valign=c, cfbox=torstenbackground]{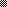}  & \includegraphics[scale=3,valign=c, cfbox=torstenbackground]{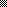} &  \includegraphics[scale=3,valign=c, cfbox=torstenbackground]{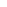}  & \includegraphics[scale=3,valign=c, cfbox=torstenbackground]{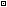} &   \includegraphics[scale=3,valign=c, cfbox=torstenbackground]{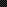}  & \includegraphics[scale=3,valign=c, cfbox=torstenbackground]{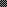} \vspace{1pt} \\[1pt] \hline
SSIM & 1 & -1     & 0      & -0.02  & 0.18   & 0.75  \\[1pt] \hline
MSE  & 0 & 64,322 & 31,642 & 32,856 & 19,764 & 5,909 \\
\end{tabular}
\end{adjustbox}
\caption{Visualization of a reference image in the first column, that is compared with the images from the first to sixth column showing the SSIM and MSE values.}
\vspace{-0.2cm}
\label{fig:ssim:background}
\end{table}
\vspace{-0.4cm}
\subsection{Threat Model}
\label{sec:problem:threatmodel}
Our threat model specifies DNN model modification scenarios the attacker can undertake with the goal of removing the watermark from the trained and watermarked model. 

\vspace{0.1cm}
\noindent \textbf{Fine-Tuning.} In fine-tuning~\cite{tajbakhsh2016finetuning,simonyan2015finetuning}, the adversary continues training with a dataset akin to the original training dataset in the hope of removing watermarks that are added on top of the main task. Here we assume, that the adversary is aware of the training procedure including hyperparameters, and hence can adopt the settings for model modifications. Specifically, the learning rate is either kept at parity with the original training or, alternatively, can be decreased from the original value. %Both fine-tuning scenarios are legitimate. 
In most benign fine-tuning scenarios, such a reduction of the learning rate is a typical approach to preserve the already trained meaningful features and only provoke small changes caused by the new dataset. Alternatively, the adversary can fine-tune the model on a different dataset, which may necessitate the substitution of the last model layer with an untrained counterpart due to a different number of label classes.

\vspace{0.1cm}
\noindent \textbf{Pruning.} Pruning~\cite{han2015pruning} is normally used to reduce the DNN size to facilitate deployment in smaller setups like embedded devices. As the adversary can arbitrarily modify model weights, parameters can be pruned in the hope of removing the watermark while keeping a reasonable main task performance. This entails the elimination of a specific proportion of parameters, called pruning level, characterized by the lowest absolute values within the model, as those parameters are deemed to have the most marginal influence on the model's overall performance. Such pruning methods can be combined with fine-tuning, which is called fine-pruning~\cite{finepruning_firstprune,finepruning_firsttune}.

\vspace{0.1cm}
\noindent \textbf{Adaptive Adversary.} An informed adversary, possessing knowledge of the watermarking methodology, may attempt to manipulate the existing watermark~\cite{katzenbeisser2016information,uchida2017embedding} leveraging the same embedding technique. Thereby, the adversary can invent a new watermark that can either contain meaningful or random data and embed the watermark in the hope of removing or replacing the original one. Removing the watermark would prevent ownership claims by the model creator. A eplacement would transfer the possibility of claiming model ownership to the adversary. Usually, the watermark is kept secret and the adversary is not aware of the watermark's data.

\section{Approach}
\label{sec:approach}
In this section, we present our general concept in \hyperref[sec:approach:overview]{\sect\ref{sec:approach:overview}}, followed by details about the generation of transposed models in \hyperref[sec:approach:transpose]{\sect\ref{sec:approach:transpose}}, the composition of the watermark in \hyperref[sec:approach:watermark]{\sect\ref{sec:approach:watermark}}, and details on the training procedure in \hyperref[sec:approach:moo]{\sect\ref{sec:approach:moo}}.

\vspace{-0.3cm}
\subsection{Overview}
\label{sec:approach:overview}
We propose \textit{\ourname}, a white-box and multi-bit\footnote{We embed data within the watermark instead of solely relying on the presence/absence of the watermark.} DNN watermarking method, that embeds a watermark secret matching to a specific watermark key within the complete set of model parameters. The watermark key has the form of a regular output vector of the DNN and the watermark secret is represented by an input-like sample of the DNN, that can contain arbitrary information, e.g., random text superimposed on an image, without necessitating visual or context-wise similarity to actual dataset samples. After legal or illegal model distribution of the trained and watermarked model, \ournameGen verification process can use the key to extract the embedded secret from the DNN and, thus, claim ownership and potentially copyright infringement. Below, we describe \ournameGen principle and outline the successive steps of \ourname during the model life-cycle.

\begin{figure}[tb]
  \centering
  \includegraphics[width=\linewidth]{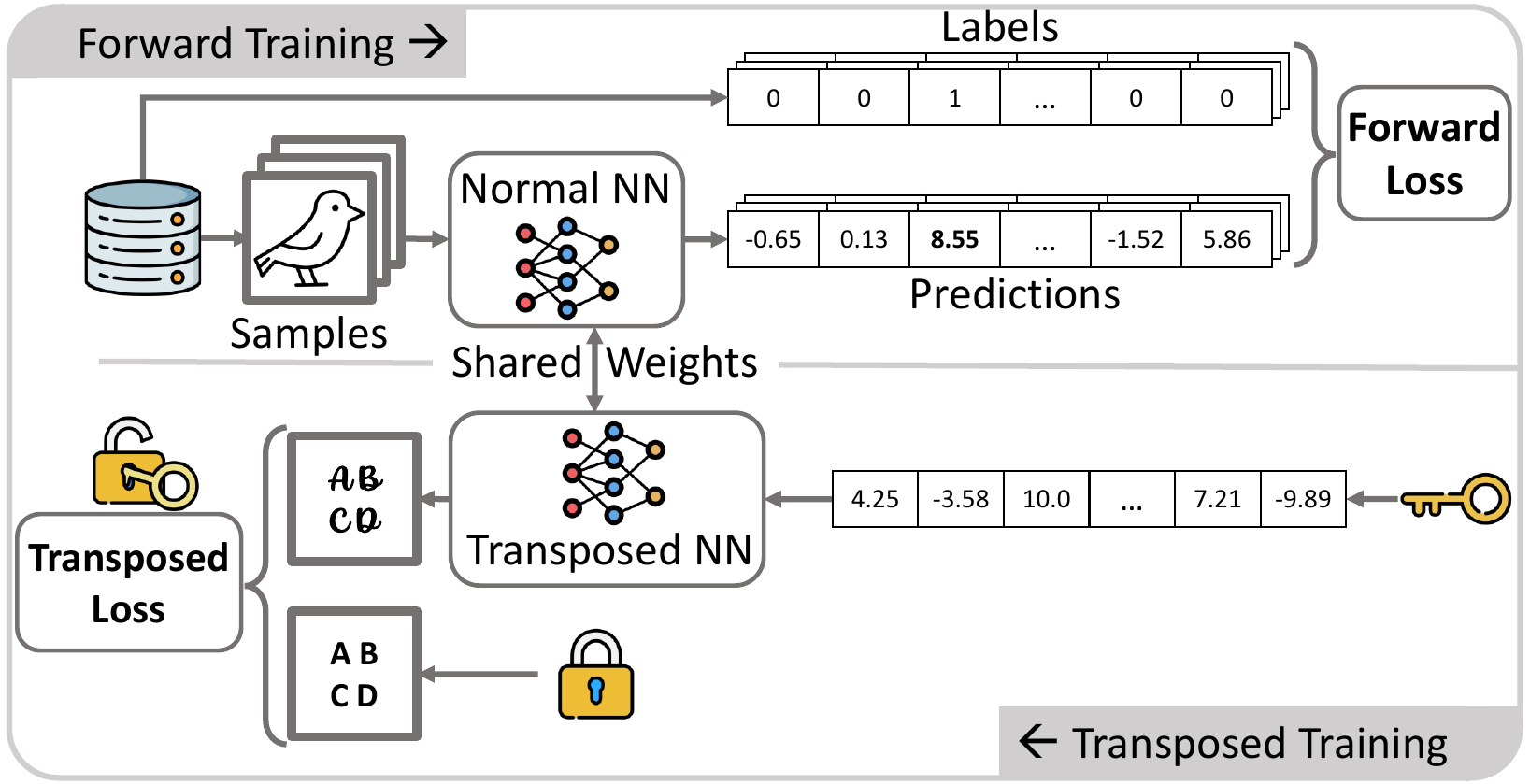}
  \caption{Visualization of main task forward training and watermark (key \& lock) embedding in transposed training.}
\vspace{-0.5cm}
\label{fig:transposed}
\end{figure}

\vspace{0.1cm}
\noindent \textbf{Principle of \ourname.} In the process of embedding a watermark into the model, \ourname employs transposed model training, as visualized in \hyperref[fig:transposed]{\fig\ref{fig:transposed}}. Therefore, we construct a transposed model architecture that is constructed using the method presented in \hyperref[sec:approach:transpose]{\sect\ref{sec:approach:transpose}}
and establish weight sharing (similar to Siamese Networks~\cite{koch2015siamese}) between the standard model and transposed model by assigning the parameters of each layer to the respective transposed layer. Consequently, these shared weights can be subject to regular training for the main task via conventional (forward) training with respect to the forward loss, e.g., cross-entropy, as illustrated in the upper portion of \hyperref[fig:transposed]{\fig\ref{fig:transposed}}. Simultaneously, we perform the transposed training with the predefined watermark key and secret visualized as key and lock in \hyperref[fig:transposed]{\fig\ref{fig:transposed}}, representing the training data. The key, a prediction-like vector, is fed into the transposed model, generating an output akin to the input data of the regular DNN, e.g., an image, as visualized in the lower part of \hyperref[fig:transposed]{\fig\ref{fig:transposed}}. During transposed training, the shared model weights are optimized with respect to a transposed loss, that is calculated by comparing the output of the transposed model with the watermark's secret. By applying both, forward and backward training, \ourname is capable of entangling the main task and the watermark within all model layers of the \dnn.

\begin{figure}[tb]
  \centering
  \includegraphics[width=\linewidth]{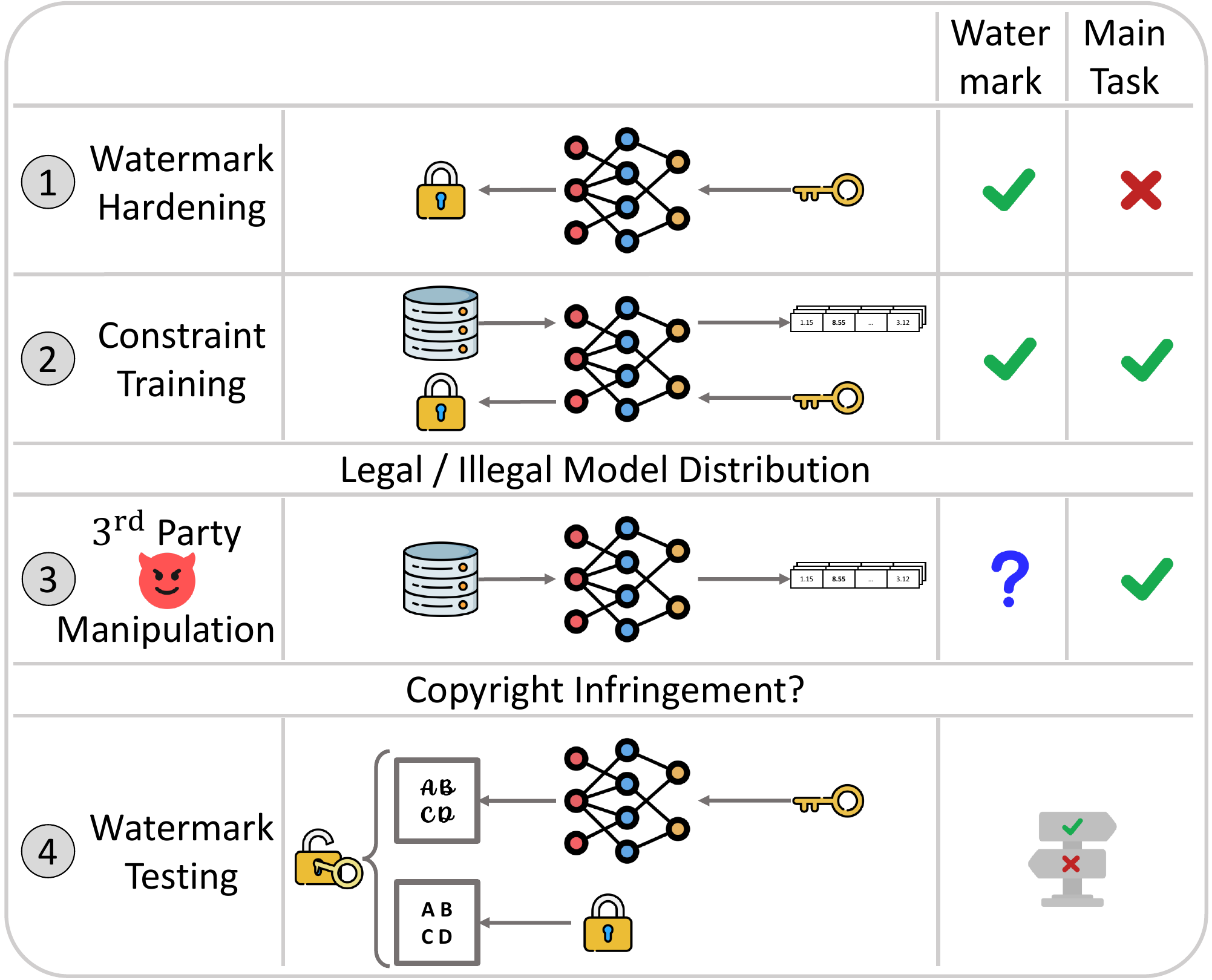}
  \caption{Overview of \ournameGen life cycle steps.}
\vspace{-0.4cm}
\label{fig:overview}
\end{figure}

\vspace{0.1cm}
\noindent \textbf{\ourname Life Cycle.} During a model's life-cycle, \ourname follows the four steps visualized in \hyperref[fig:overview]{\fig\ref{fig:overview}}. The untrained model is initialized with random parameters, that neither performs well on the main task for forward model inference nor on the watermark for transposed inference. 1)~In the watermark hardening phase, we initialize the parameters of the model by transposed training on the watermark until a self-defined sufficient enough watermark quality is reached, which essentially means, that the model is overfitted on the watermark\footnote{We suggest an SSIM value above 0.95 as the watermark quality threshold, as such values can be achieved fast in transposed-only training since the watermark consists of a limited small amount of key-value pairs. In our experiments, we combine the SSIM threshold with a maximum of 10,000 learning epochs.}. Thereby, the watermark builds the basis for the main task in the consecutive forward training within the model parameters. 2)~During the constraint training phase, normal forward model training is equipped such that the already embedded watermark from step 1 persists. Thereby, we alternate between optimizing the two tasks, which is described in detail in~\hyperref[sec:approach:moo]{\sect\ref{sec:approach:moo}}. The training can mainly focus on optimizing the main task, as the effect of the watermark task during optimization is minimal, due to the foregone watermark hardening step. The parameters are prepared during step 1 such that the watermarking task yields a negligible loss compared to the main task and essentially functions as a constraint during normal model training. 3)~After model distribution, the model is manipulated by a third party, e.g., fine-tuned\footnote{Step 3 can differ depending on the scenario or the attack. We describe various scenarios in~\hyperref[sec:problem:threatmodel]{\sect\ref{sec:problem:threatmodel}} and evaluate \ourname against them in \hyperref[eval:modelmanipulations]{\sect\ref{eval:modelmanipulations}} and \hyperref[eval:adaptive]{\sect\ref{eval:adaptive}}.}. As long as the main task performance is preserved, the watermark should remain embedded. 4)~Finally, a transposed inference on the watermark key is conducted to extract the watermark data, which is verified against the ground truth watermark secret by human-only inspection or machines. 
Since the main task relies on the fortified parameters established in step 1, creating an inherent interconnection between the tasks, substantial modifications made to the watermark in step 3 directly influence the main task. This direct impact enhances the overall robustness of the watermark. 

\vspace{0.1cm}
\noindent To enable \ourname, we must define several components: 
First, in \hyperref[sec:approach:transpose]{\sect\ref{sec:approach:transpose}}, we define rules for the creation of a transposed model from a given model architecture. Second, we specify how the key and the secret of the watermark are composed in \hyperref[sec:approach:watermark]{\sect\ref{sec:approach:watermark}}. Third, in \hyperref[sec:approach:moo]{\sect\ref{sec:approach:moo}}, we determine how to train while maintaining the watermark.

\vspace{-0.3cm}
\subsection{Transposed Model Generation}
\label{sec:approach:transpose}
\vspace{-0.1cm}
To ensure fulfillment of the generalizability requirement from \hyperref[sec:problem:objectives]{\sect\ref{sec:problem:objectives}}, we establish guidelines for the generation of transposed models. Thereby, we define how model layers and connections are translated to the transposed version\footnote{Please note that the transposed model can be generated solely from the weights of the original model and does not require any additional parameters.}. Linear layers and batch normalization layers~\cite{batchnorm} are straightforward mathematical operations and easy to transpose, as discussed in \hyperref[sec:background]{\sect\ref{sec:background:transposed}}. For pooling layers~\cite{pooling}, we leverage interpolation based on the nearest-neighbor algorithm~\cite{Rukundo2012_nearestneighborinterpolation}. Convolutional layers~\cite{convolution} are transposed with deconvolutions\footnote{We adapt the settings, e.g., kernel size and stride, from the convolution with potential adjustments to padding to ensure the output of the deconvolution matches the original input.} as in~\cite{zeiler2010deconvolutional}. Dropout layers~\cite{dropout} and activation functions, e.g., ReLU~\cite{agarap2018deep_relu}, are used likewise to their untransposed functionality.

\vspace{0.1cm}
\noindent \textbf{Skip Connections.} Skip connections fork the data processing within the model architecture and merge the data from both branches at a later stage, alleviating the vanishing gradient problem and improving the accuracy of DNNs. Skip connections effectively perform an operation akin to $a + b = c$ during the merging process within the forward path. Hence, they are difficult or impossible to reverse as only the output $c$ is provided during transposed training, rendering $a$ and $b$ indistinct. To transpose the skip connection, we freeze one part of the connection, such as $b$, during transposed training. Thus, by the inverse of the mathematical operation between $a$ and $b$ the skip connection can be transposed utilizing $c$ and the frozen $b$. This effectively adds $b$ to the watermark's key. To get a reasonable estimation of a realistic value for $b$, we start training an unwatermarked model for a few initial epochs.

\vspace{0.1cm}
\noindent \textbf{Additional Dropout Layers.} To ensure the robustness requirement from \hyperref[sec:problem:objectives]{\sect\ref{sec:problem:objectives}}, \ourname strives for a robust entanglement between the watermark and the main task. Therefore, the watermark needs to be embedded within all model layers during the watermark hardening step. To facilitate the entanglement between the watermark and the main task, spreading the watermark across multiple parameters must be enforced. Such a behavior can be achieved within model architectures by utilizing dropout layers~\cite{dropout}. For model architectures that lack inherent dropout layers, we artificially add such layers into the transposed architecture. Specifically, dropout layers are incorporated after each convolutional and linear layer, with the exception of the final layer responsible for producing the ultimate output of the transposed model. The dropout rate is an insensitive parameter, that must be set to some reasonable value, which can be quickly identified by analyzing the first few update steps during watermark hardening. Higher dropout rates extend the duration of the hardening process but do not compromise \ournameGen functionality.

\subsection{Watermark Composition}
\label{sec:approach:watermark}
\noindent \textbf{Watermark Key.} The key's structure must align with the dimensions of a regular output vector of the (forward) model. Generally, there are no constraints on the values within this vector, allowing for arbitrary and extreme values, that are usually not encountered in forward prediction vectors. However, we need to enforce an overlap of the model's forward output value range and the watermark key's value range. Otherwise, e.g., if the key only consists of positive values,  the model will most likely group the outputs of the main tasks to different value ranges, e.g., negative values. Such a separation prevents tight entanglement of the two tasks and encourages the model to handle the tasks in a multi-task instead of a constraint-task manner. As a result of separated value ranges, the watermark can be removed from the model with minimal effects on the main task, contrary to scenarios where the value ranges overlap. To address this, we generate random key vectors with values between a predefined range from -10 to 10, as visualized in \hyperref[fig:transposed]{\fig\ref{fig:transposed}}, given that most random initialized models predominantly generate values around zero.

\vspace{0.1cm}
\noindent \textbf{Watermark Secret.} The watermark's secret, adaptable to regular input sample dimensions like images, can be 1-bit using a random image or multi-bit with additional content like text. Similar to the watermark key values, the values of the watermark secret should fall within the range of typical input data. This ensures an intertwined relationship between the watermark and the main task parameters, ultimately enhancing the robustness of the watermark. When employing a loss function solely based on structural similarity between images, there might be challenges in producing outputs within the desired range.  To address output range challenges, we employ a dual-loss strategy in transposed training. SSIM ensures structural similarity with the watermark secret, while MSE maintains exact secret values in the output.

\vspace{0.1cm}
\noindent \textbf{Multi-Key.} Employing not just one, but multiple unique key-secret pairs within a single watermark significantly increases the capacity of the multi-bit watermark addressing the capacity requirement from \hyperref[sec:problem:objectives]{\sect\ref{sec:problem:objectives}}. This approach allows for the embedding of a greater volume of information. Furthermore, it enhances the robustness of the watermark as multiple keys influence a larger portion of the model's parameters, complicating erasure attempts by third parties. These distinct key-secret pairs are inputted into the transposed model as a unified batch. This method compels the model to grasp the underlying structure of the secrets and integrate the watermark throughout all layers. Additionally, this approach offers the advantage of preventing the parameters in the transposed model from memorizing specific key-secret pairs. Instead, the model generalizes the functionality underlying the samples, reinforcing learned capabilities across various layers. To optimize this approach, it is advisable to ensure that the different key-secret pairs share a consistent structure, such as all containing textual information within images. This uniformity enhances the model's ability to learn and embed diverse information effectively.

\begin{table*}[]
\centering
\fontsize{7pt}{8pt}\selectfont
  \caption{In these experiments \mnist~\cite{mnist} was trained on a CNN with a learning rate of 0.001 for five epochs.}
  \label{tab:generalfunc}
  \begin{tabular}{l|c|c|c|c|c|c|c|c|c|c|c|c|c}
    & \multirow{3}{*}{\shortstack{Number \\ of keys}}& \multicolumn{2}{c|}{Untrained} & \multicolumn{3}{c|}{Watermark}&\multicolumn{3}{c|}{\multirow{2}{*}{Training}} & \multirow{3}{*}{\shortstack{Extracted \\ Watermark \\ Figure}} & \multirow{3}{*}{\shortstack{Watermark\\Considered\\Valid}}& \multicolumn{2}{c}{4 epochs fine-tuning with } \\
    & & \multicolumn{2}{c|}{Model}& \multicolumn{3}{c|}{Hardening}&\multicolumn{1}{c}{}&\multicolumn{1}{c}{}&  &  & & \multicolumn{2}{c}{\nicefrac{1}{10} of training learning rate}\\
    \cline{3-10}
    \cline{13-14}
    & & Accuracy & SSIM & Steps & Accuracy & SSIM &Constraint&Accuracy&SSIM  &  &&  Accuracy & SSIM\\
    \hline
    \hline
    (1) &-& 10.22\% & 0.00 & - & - & - &x& 89.88\% & -0.06&\hyperref[fig:1:exp2:a]{\fig\ref{fig:1:exp2:a}}&x  & - & - \\
    (2)  & 1&10.22\% & 0.00 & 7,000& 8.57\%& 0.95 &-& -& -&\hyperref[fig:1:exp1:c]{\fig\ref{fig:1:exp1:c}}&\checkmark  & - & - \\
    (3)  & 1&10.22\% & 0.00 & 7,000& 8.57\%& 0.95 &\checkmark& 88.37\%& 0.95&\hyperref[fig:1:exp3:a]{\fig\ref{fig:1:exp3:a}}&\checkmark  & 92.73\% & 0.95 \\
    (4)  & 10&10.22\% & 0.00 & 10,000& 8.92\%& 0.93 &\checkmark& 87.69\%& 0.91&\hyperref[fig:1:exp4:b]{\fig\ref{fig:1:exp4:b}} &\checkmark  & 89.73\% & 0.93 \\
    (5)  & 11&10.22\% & 0.00 & 10,000& 8.88\%& 0.92 &\checkmark& 89.10\%& 0.91&\hyperref[fig:1:exp4:a]{\fig\ref{fig:1:exp4:a}}&\checkmark  & 89.86\% & 0.93 \\
    
\end{tabular}
\vspace{-0.2cm}
\end{table*}
\vspace{-0.4cm}
\subsection{Constraint Training}
\label{sec:approach:moo}
Watermarking the model in the proposed way (\cf~\hyperref[fig:transposed]{\fig\ref{fig:transposed}}) essentially corresponds to simultaneously learning two separate tasks within one model, resulting in a multi-objective optimization problem consisting of the model's general functionality as task one and the watermark in the transposed model as the second task. After overfitting the transposed model to the watermark in the watermark hardening phase (step 1 of~\hyperref[fig:overview]{\fig\ref{fig:overview}}), the watermarking task can be considered as a constraint to the main task in step 2 in~\hyperref[fig:overview]{\fig\ref{fig:overview}}. This entails, that we optimize the main task while keeping the watermark functionality. To execute this optimization, we leverage sequential optimization, essentially alternating between optimizing the model parameters for the main task and the watermark. Alternatives to this optimization approach are discussed in \hyperref[sec:discussion:optimization]{\sect\ref{sec:discussion:optimization}}.

\section{Evaluation}
\label{sec:eval}
\noindent \textbf{Hardware \& Experimental Setup.} Experiments are implemented in PyTorch, a prominent Python-based machine learning library~\cite{pytorch, paszke2019pytorch,van1995python}, on a server featuring an AMD EPYC 7413 24-Core Processor (64-bit) with 96 processing units and 128GB main memory. An NVIDIA A16 GPU with 4 virtual GPUs, each equipped with 16GB GDDR6 memory, is accessible via CUDA~\cite{cuda}.

\vspace{0.1cm}
\noindent \textbf{Datasets \& Model Architectures.} We use common datasets mainly focusing on image classification with \mnist~\cite{mnist}, \cifar~\cite{cifar}, \gtsrb~\cite{gtsrb}, and \cifarhundred~\cite{cifar} trained on models of different types and sizes, namely CNNs (with and without batch normalization), \resneteighteen, \resnetthirtyfour~\cite{resnet}, ViT~\cite{vit}, and VGG11~\cite{vgg_eleven}.\footnote{We use PyTorch model instances for predefined model architectures.}

\vspace{0.1cm}
\noindent \textbf{Default Scenario.} Throughout our experiments, we systematically vary model architecture, dataset, and hyperparameters to illustrate the versatility of our approach. Unless otherwise specified, our default scenario uses \mnist~\cite{mnist} trained on a CNN consisting of two convolution layers both followed by a ReLU and a 2D max pooling layer and followed by three fully connected layers of decreasing output sizes  (512, 256, 10). For training purposes, we employ separate Adam optimizers with a learning rate of 0.0001, both for the primary task and transposed training. We trained the model for five epochs.

\vspace{-0.2cm}
\subsection{General Functionality}

\begin{figure}[tb]
\centering
\begin{subfigure}{0.13\linewidth}
    \includegraphics[width=\textwidth, cfbox=torstenbackground]{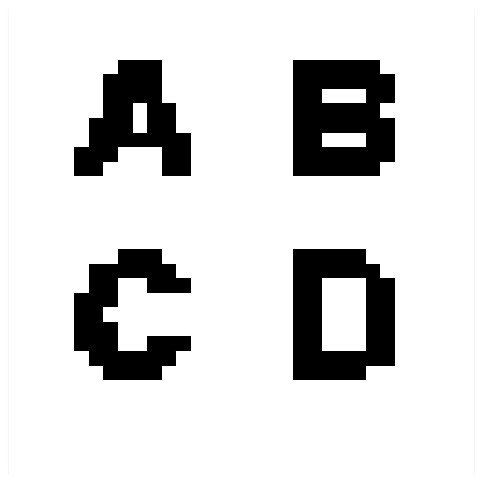} % 1_only_wm_final
    \vspace{-0.4cm}
\caption{}
    \label{fig:1:exp1:a}
\end{subfigure}
\hfill
\begin{subfigure}{0.13\linewidth}
    \includegraphics[width=\textwidth, cfbox=torstenbackground]{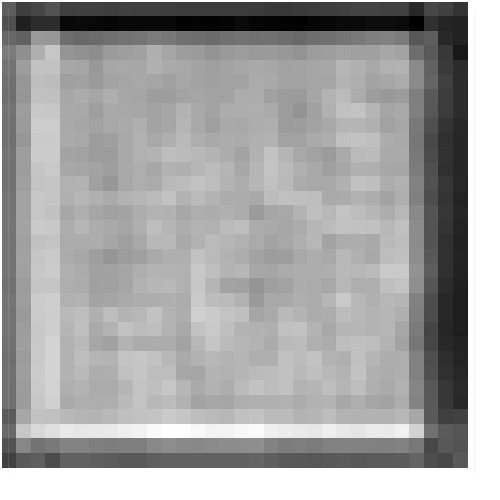} % 1_only_wm_final
    \vspace{-0.4cm}
\caption{}
    \label{fig:1:exp1:b}
\end{subfigure}
\hfill
\begin{subfigure}{0.13\linewidth}
    \includegraphics[width=\textwidth, cfbox=torstenbackground]{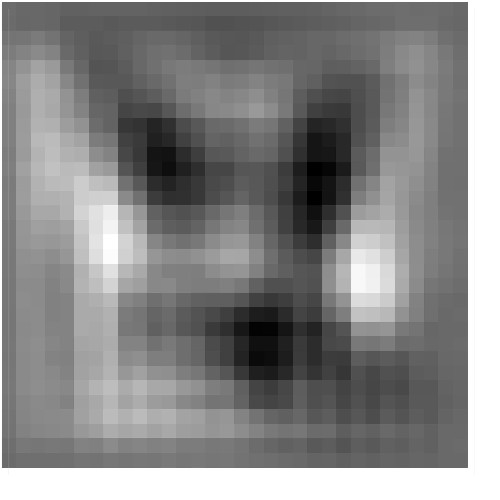} % 2_model_plain_final
    \vspace{-0.4cm}
\caption{}
    \label{fig:1:exp2:a}
\end{subfigure}
\hfill
\begin{subfigure}{0.13\linewidth}
    \includegraphics[width=\textwidth, cfbox=torstenbackground]{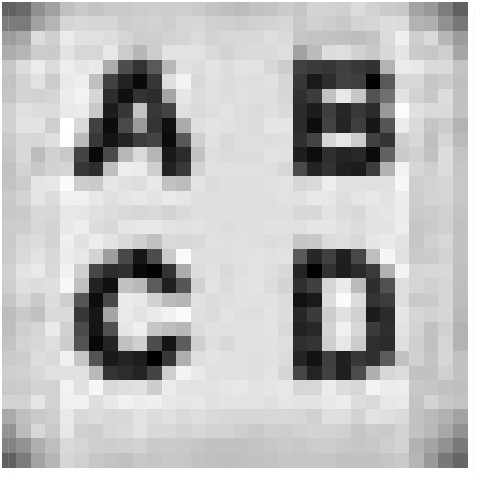} % 1_only_wm_final
    \vspace{-0.4cm}
\caption{}
    \label{fig:1:exp1:c}
\end{subfigure}
\hfill
\begin{subfigure}{0.13\linewidth}
    \includegraphics[width=\textwidth, cfbox=torstenbackground]{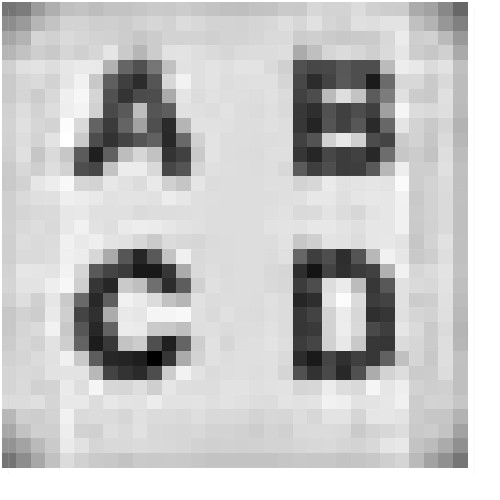} % 3_model_with_wm
    \vspace{-0.4cm}
\caption{}
    \label{fig:1:exp3:a}
\end{subfigure}
\hfill
\begin{subfigure}{0.13\linewidth}
    \includegraphics[width=\textwidth, cfbox=torstenbackground]{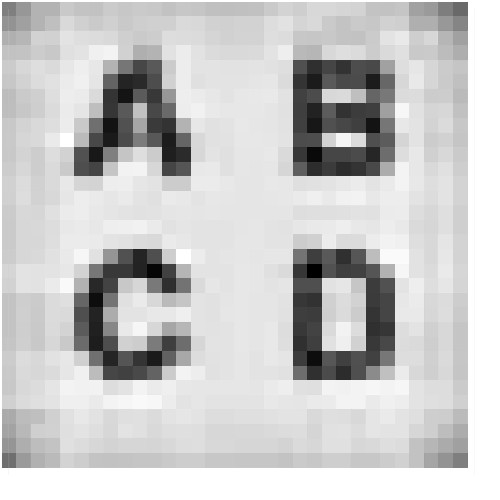} % 4_multikeys_11
    \vspace{-0.4cm}
\caption{}
    \label{fig:1:exp4:b}
\end{subfigure}
\hfill
\begin{subfigure}{0.13\linewidth}
    \includegraphics[width=\textwidth, cfbox=torstenbackground]{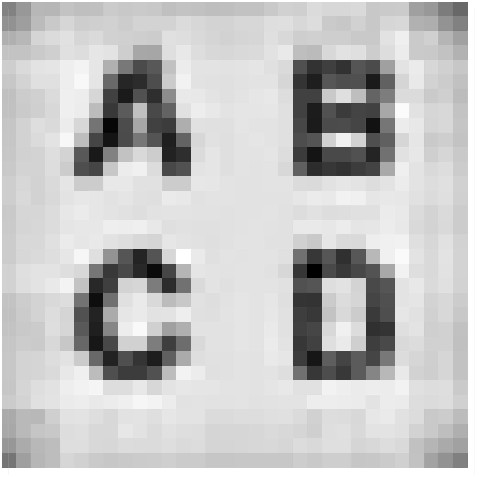} % 4_multikeys_11
    \vspace{-0.4cm}
\caption{}
    \label{fig:1:exp4:a}
\end{subfigure}
\caption{Visualization of (a) the watermark secret and (b-g) extracted watermarks for the experiments listed in \hyperref[tab:generalfunc]{\tab\ref{tab:generalfunc}}.}
\vspace{-0.4cm}
\label{fig:1}
\end{figure}

\noindent \textbf{Watermark Definition.} A watermark for \ourname consists of one or multiple watermark keys and secrets, which are kept confidential. In our experiments, the keys are vectors consisting of ten randomly chosen values between -10 and 10\footnote{The precise random key vector for the experiments with one key is provided in \hyperref[app:additionaldetails]{\app\ref{app:additionaldetails}}.}. The chosen secrets are images containing four letters like ``ABCD'', as visualized in \hyperref[fig:1:exp1:a]{\fig\ref{fig:1:exp1:a}} and \hyperref[fig:secrets_default]{\appFig\ref{fig:secrets_default}}.

\vspace{0.1cm}
\noindent \textbf{Baseline - No Watermark.} First, we trained a model without a watermark, which serves as a baseline for model performance. As can be seen in (1) in \hyperref[tab:generalfunc]{\tab\ref{tab:generalfunc}}, we reached a model accuracy of 89.88\%. When trying to extract a watermark before and after training, we get images as in \hyperref[fig:1:exp1:b]{\fig\ref{fig:1:exp1:b}} and \hyperref[fig:1:exp2:a]{\fig\ref{fig:1:exp2:a}}, respectively. The images clearly show no relation or similarity to \hyperref[fig:1:exp1:a]{\fig\ref{fig:1:exp1:a}}, indicating the absence of the watermark essentially fulfilling the integrity requirement from \hyperref[sec:problem:objectives]{\sect\ref{sec:problem:objectives}}.
Additionally, the two pictures yield an SSIM of 0.00 and -0.06 when being compared to \hyperref[fig:1:exp1:a]{\fig\ref{fig:1:exp1:a}}, confirming that watermarked and unwatermarked models are clearly distinguishable. However, human perception instead of a low SSIM should be the main criterion for the decision, as even for small SSIMs close to zero human observers can still recognize similarities between images (\cf~\hyperref[fig:ssim:background]{\tab\ref{fig:ssim:background}} in \hyperref[sec:background:ssim]{\sect\ref{sec:background:ssim}}).

\vspace{0.1cm}
\noindent \textbf{Watermark Hardening.} Next, in step 1 of \ourname (\cf~\hyperref[fig:overview]{\fig\ref{fig:overview}}) we perform transposed training as described in \hyperref[sec:approach:overview]{\sect\ref{sec:approach:overview}} to embedded a watermark consisting of one key-secret pair into an untrained model. As transposed loss, we combine SSIM and MSE between the transposed model output and the watermarks ground truth secret. We name each adjustment of the transposed model parameters by the optimizer as a hardening step. As presented in (2) in \hyperref[tab:generalfunc]{\tab\ref{tab:generalfunc}} we reached an SSIM of 0.95 after 7,000 hardening steps. The resulting image after watermark extraction depicted in \hyperref[fig:1:exp1:c]{\fig\ref{fig:1:exp1:c}} clearly shows the content of the ground truth secret \hyperref[fig:1:exp1:a]{\fig\ref{fig:1:exp1:a}} and an existing watermark can be attested. As expected, the accuracy on the main task remained \naive with 10.22\% and 8.57\% accuracy before and after training, respectively.

\vspace{0.1cm}
\noindent \textbf{Constraint Training.} In \ournameGen second step, we train the model's main task while keeping the watermark embedded as described in~\hyperref[sec:approach:moo]{\sect\ref{sec:approach:moo}}. As reported in (3) in \hyperref[tab:generalfunc]{\tab\ref{tab:generalfunc}}, the watermark remains embedded, while the main task accuracy of 88.37\% is achieved, fulfilling the reliability requirement from \hyperref[sec:problem:objectives]{\sect\ref{sec:problem:objectives}}. Hence, we observe a negligible accuracy drop compared to the unwatermarked model ((2) in \hyperref[tab:generalfunc]{\tab\ref{tab:generalfunc}}) satisfying the fidelity requirement from \hyperref[sec:problem:objectives]{\sect\ref{sec:problem:objectives}}. To emphasize this important fact, we visualize the main task loss for unwatermarked and watermarked training in \hyperref[fig:mainlosses]{\appFig\ref{fig:mainlosses}} showing minimal differences. These results could be reproduced independently of the optimizer used for the different tasks, which we elaborate on in \hyperref[app:additionalexperiments]{\appSect\ref{app:additionalexperiments}}.

\vspace{0.1cm}
\noindent \textbf{Multi-Key.} Next, we investigate multiple watermark key-secret pairs, as explained in \hyperref[sec:approach:watermark]{\sect\ref{sec:approach:watermark}}. We employed ten and eleven keys to show the independence from the number of classes in the dataset, ten for \mnist~\cite{mnist}. The secrets are distinct four-character images\footnote{During visualizations, we stick to the first image containing ``ABCD''.}, as visualized in \hyperref[fig:secrets_default]{\appFig\ref{fig:secrets_default}}. The results for ten and eleven keys are shown in (4) and (5) in \hyperref[tab:generalfunc]{\tab\ref{tab:generalfunc}}, respectively, and show that \ourname embeds the watermark successfully. Thus, we can increase the watermark capacity without a significant negative impact on the watermark's or the model's performance\footnote{We observe a slight increase in main-task accuracy for (5) in \hyperref[tab:generalfunc]{\tab\ref{tab:generalfunc}}, indicating, that the watermarking is tightly enmeshed with the main task and serves as regularization in this experiment. However, the general observation is a minimal drop in accuracy due to watermark embedding.}. Notably, we stopped hardening after 10,000 hardening steps but it would be possible to continue training until a 0.99 SSIM is reached\footnote{We report the mean SSIM values for multiple keys. If not specifically mentioned, the means do not contain extreme outliers.}. While it is possible to increase the number of keys and thus the capacity, which we discuss in \hyperref[eval:capacity]{\sect\ref{eval:capacity}}, we proceed with eleven keys to showcase the functionality of \ourname. To show the independence from the concrete secret images, we also conducted the same experiments with other images, visualized in \hyperref[fig:secrets_airplane]{\fig\ref{fig:secrets_airplane}} and \hyperref[fig:secrets_airplane_real]{\fig\ref{fig:secrets_airplane_real}}, yielding similar results.

\subsection{Model Manipulations}
\label{eval:modelmanipulations}
Next, we evaluate illegal and legal third-party model manipulations, that are applied in step 3 of \hyperref[fig:overview]{\fig\ref{fig:overview}}.

\begin{figure}[tb]
\centering
\begin{subfigure}{0.13\linewidth}
    \includegraphics[width=\textwidth, cfbox=torstenbackground]{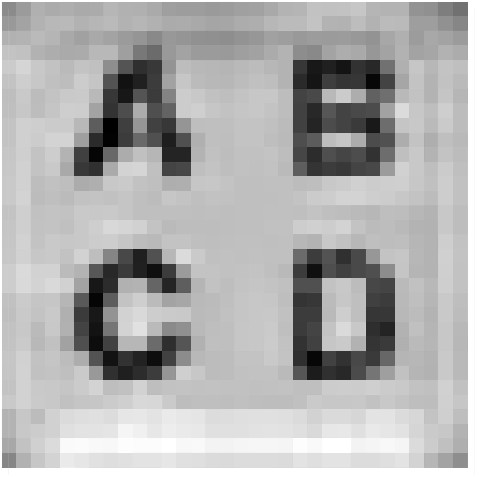} % 9_default_final
    \vspace{-0.4cm}
\caption{}
    \label{fig:2:exp6:a}
\end{subfigure}
\begin{subfigure}{0.13\linewidth}
    \includegraphics[width=\textwidth, cfbox=torstenbackground]{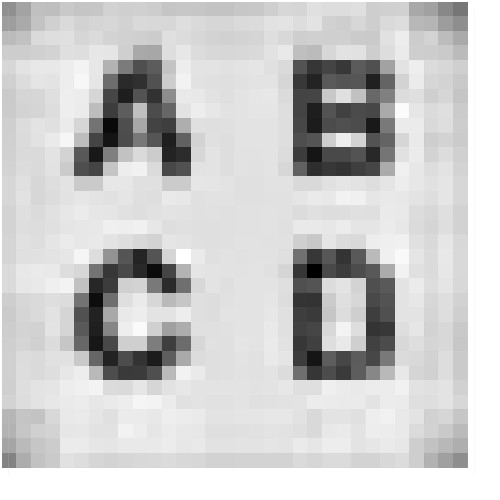} % 9_default_final
    \vspace{-0.4cm}
\caption{}
    \label{fig:2:exp6:b}
\end{subfigure}
\begin{subfigure}{0.13\linewidth}
    \includegraphics[width=\textwidth, cfbox=torstenbackground]{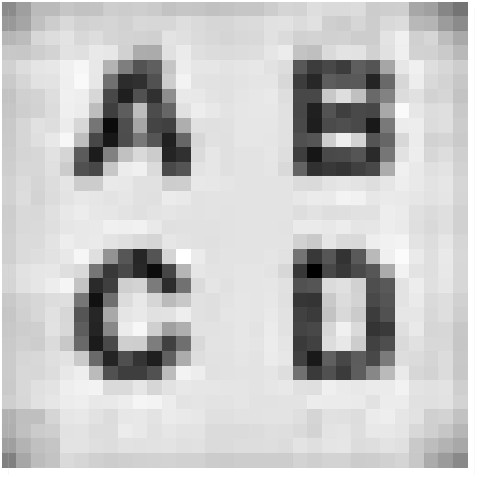} % 10_finetune_on_testdataset
    \vspace{-0.4cm}
\caption{}
    \label{fig:2:exp7:a}
\end{subfigure}
\hfill
\begin{subfigure}{0.13\linewidth}
    \includegraphics[width=\textwidth, cfbox=torstenbackground]{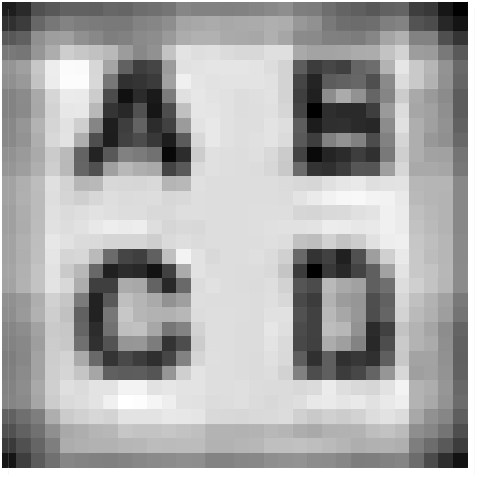} % 11_change_last_layer
    \vspace{-0.4cm}
\caption{}
    \label{fig:2:exp8:a}
\end{subfigure}
\hfill
\begin{subfigure}{0.13\linewidth}
    \includegraphics[width=\textwidth, cfbox=torstenbackground]{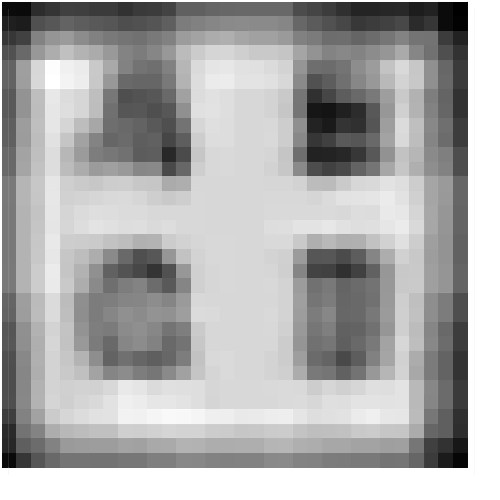} % 11_change_lastlayer_on_plain_model
    \vspace{-0.4cm}
\caption{}
    \label{fig:2:exp8:b}
\end{subfigure}
\hfill
\begin{subfigure}{0.13\linewidth}
    \includegraphics[width=\textwidth, cfbox=torstenbackground]{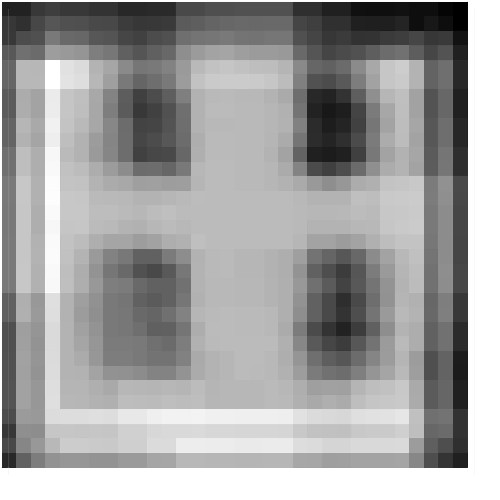} % 16_pruning_03
    \vspace{-0.4cm}
\caption{}
    \label{fig:2:exp8:c}
\end{subfigure}
\hfill
\begin{subfigure}{0.13\linewidth}
    \includegraphics[width=\textwidth, cfbox=torstenbackground]{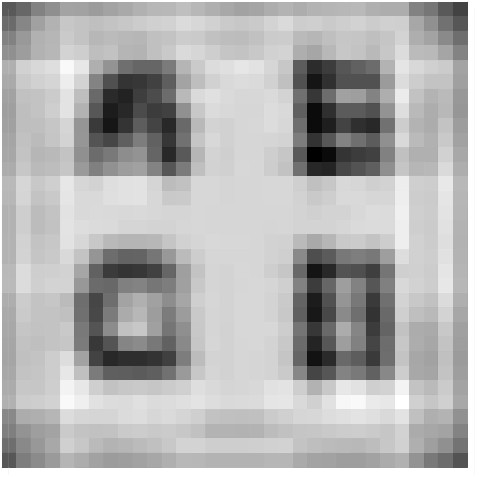} % 16_pruning_05
    \vspace{-0.4cm}
\caption{}
    \label{fig:2:exp8:d}
\end{subfigure}
\caption{Visualization of extracted watermarks for the fine-tuning experiments listed in \hyperref[tab:finetune]{\tab\ref{tab:finetune}} in (a-c).
Figures (d-f) visualize pruning with 60\%, 80\%, and 90\% respectively, while (g) shows fine-pruning with 40\%.}
\vspace{-0.4cm}
\label{fig:2}
\end{figure}

\vspace{0.1cm}
\noindent \textbf{Fine-Tuning.} To evaluate \ournameGen robustness against fine-tuning, as described in \hyperref[sec:problem:threatmodel]{\sect\ref{sec:problem:threatmodel}}, we continued training on the \mnist train set after executing our default scenario for another two epochs (half of the original five epochs rounded down) with the same learning rate, as well as with \nicefrac{1}{10} of the original learning rate (similar to~\cite{uchida2017embedding,chen2019deepmarks,xie2021deepmark,lemerrer2019frontierstitching,adi2018turning,li2021spreadtransform}). After fine-tuning for two epochs, we continued for another two epochs with identical settings to showcase \ournameGen behavior under excessive fine-tuning conditions. To evaluate the watermark robustness against unseen data, we executed the same fine-tuning process but employed the \mnist~\cite{mnist} test set. As \hyperref[tab:finetune]{\tab\ref{tab:finetune}} shows, \ourname shows strong robustness against fine-tuning as the watermark remained embedded yielding high SSIM values and clear images (\hyperref[fig:2:exp6:a]{\fig\ref{fig:2:exp6:a}} to \hyperref[fig:2:exp7:a]{\fig\ref{fig:2:exp7:a}})\footnote{The accuracies show, that fine-tuning with \nicefrac{1}{10} of the original learning rate is a better setting if an adversary wants to increase the model accuracy on a third-party dataset.}. Later, in \hyperref[sec:eval:appindependence]{\sect\ref{sec:eval:appindependence}}, we also evaluate cross-dataset fine-tuning. In scenario (2) in \hyperref[tab:finetune]{\tab\ref{tab:finetune}} we can observe a slight increase in SSIM after 4 epochs compared to 2 epochs, which is counter-intuitive. We believe that caused by the overlapping value ranges of our watermark and the entanglement of parameters, weight changes for the forward path can have small positive effects on the watermark's similarity score.

\begin{table}
\centering
\fontsize{7pt}{8pt}\selectfont
  \caption{Fine-tuning experiments for a CNN trained using \mnist~\cite{mnist} test set with a learning rate of 0.001 for five epochs with eleven watermark keys.}
  \label{tab:finetune}
  \begin{tabular}{c|c|c|c|c|c|c}
    \multirow{2}{*}{\shortstack{Fine-Tuning \\ Scenario}}& \multicolumn{2}{c|}{2 Epochs} & \multicolumn{2}{c|}{4 Epochs}&\multicolumn{2}{c}{Wateermark}\\
    \cline{2-7}
    &ACC&SSIM&ACC& SSIM & Figure & Valid\\
    \hline
    \hline 
    (1)&88.64\%&0.90&  87.42\% & 0.88  & \hyperref[fig:2:exp6:a]{\fig\ref{fig:2:exp6:a}} &\checkmark\\
    (2)&88.56\%&0.92&  89.86\% & 0.93  & \hyperref[fig:2:exp6:b]{\fig\ref{fig:2:exp6:b}} &\checkmark\\
    (3)&96.31\%&0.92&  97.02\% & 0.92  & \hyperref[fig:2:exp7:a]{\fig\ref{fig:2:exp7:a}} &\checkmark\\
    \hline 
    \multicolumn{7}{l}{(1) Same as training learning rate (0.001) \& same data}\\
    \multicolumn{7}{l}{(2) \nicefrac{1}{10} of training learning rate (0.0001) \& same data}\\
    \multicolumn{7}{l}{(2) \nicefrac{1}{10} of training learning rate (0.0001) \& unseen data }
    
\end{tabular}
\vspace{-0.2cm}
\end{table}

\begin{figure}[t]
\centering
\includegraphics[width=0.4\textwidth]{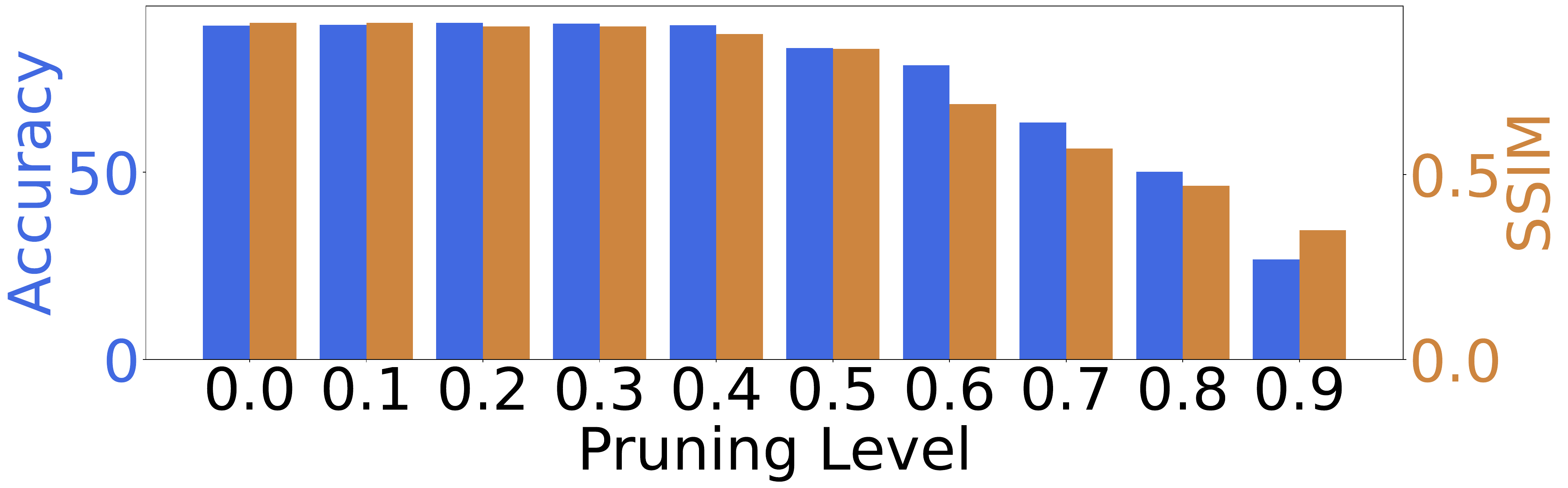}
\caption{Model accuracy on the main task and corresponding watermark SSIM for different pruning levels between 0 (unpruned) and 90\%.}
\vspace{-0.2cm}
\label{fig:pruning}
\end{figure}

\vspace{0.1cm}
\noindent \textbf{Pruning.} Besides fine-tuning, we investigate model pruning (\cf~\hyperref[sec:problem:threatmodel]{\sect\ref{sec:problem:threatmodel}}) similar to~\cite{uchida2017embedding,rouhani2019deepsigns,chen2019deepmarks,xie2021deepmark,lemerrer2019frontierstitching,li2021spreadtransform}. As depicted in \hyperref[fig:pruning]{\fig\ref{fig:pruning}}, the watermarking withstands pruning and is coupled to the main task accuracy, as for low pruning levels, which maintain the accuracy, the SSIM remains high. For example, for 60\% pruning with an accuracy drop from 89.1\% to 78.56\%, the SSIM is still 0.69, yielding \hyperref[fig:2:exp8:a]{\fig\ref{fig:2:exp8:a}}. Even for 80\% pruning, which already suffers in accuracy with 50.1\% we obtain an SSIM of 0.47 resulting in \hyperref[fig:2:exp8:b]{\fig\ref{fig:2:exp8:b}}, which is still sufficient for a human observer to identify the watermark when being aware of the ground truth secret \hyperref[fig:1:exp1:a]{\fig\ref{fig:1:exp1:a}}. Starting from 90\% pruning (\cf~\hyperref[fig:2:exp8:c]{\fig\ref{fig:2:exp8:c}}), the watermark cannot be clearly identified, but the model already decreased to 26.76\% accuracy essentially being useless. \hyperref[fig:2:exp8:d]{\fig\ref{fig:2:exp8:d}} shows the result with an SSIM of 0.62 and an accuracy of 89.79\% after two fine-tuning epochs followed by pruning with 40\%, typically called fine-pruning~\cite{finepruning_firsttune,jia2021entangledwm}. As the images prior to 90\% pruning in \hyperref[fig:2:exp8:c]{\fig\ref{fig:2:exp8:c}} are clearly distinguishable from \hyperref[fig:1:exp2:a]{\fig\ref{fig:1:exp2:a}} while \hyperref[fig:2:exp8:c]{\fig\ref{fig:2:exp8:c}} suffers low accuracy, we can conclude that \ourname is robust against model pruning essentially addressing the security requirement from \hyperref[sec:problem:objectives]{\sect\ref{sec:problem:objectives}}.

\begin{table*}
\centering
\fontsize{7pt}{8pt}\selectfont
  \caption{Experiments showing the independence of \ourname from datasets and model architectures. Tests were conducted with eleven watermark keys and fine-tuning was performed for the same number of epochs as training epochs.}
  \label{tab:datasetmodel}
  \begin{tabular}{l|c|c|c|c|c|c|c|c|c|c|c}
    & \multicolumn{2}{c|}{Untrained} & \multicolumn{3}{c|}{Watermark}&\multicolumn{2}{c|}{Constraint} & \multicolumn{2}{c|}{Fine-Tuning with} & \multirow{3}{*}{\shortstack{Extracted \\ Watermark \\ Figure}} & \multirow{3}{*}{\shortstack{Watermark\\Considered\\Valid}}\\
    & \multicolumn{2}{c|}{Model}& \multicolumn{3}{c|}{Hardening}&\multicolumn{2}{c|}{Training} & \multicolumn{2}{c|}{\nicefrac{1}{10} of training learning rate (LR)} &  &\\
    \cline{2-10}
    & Accuracy & SSIM & Steps & Accuracy & SSIM &Accuracy&SSIM&  Accuracy & SSIM  &  &\\
    \hline
    \hline
    (1)  &10.01\% & 0.00 & 10,000& 10.07\%& 0.92 & 44.16\%& 0.93 & 45.87\% & 0.93 &\hyperref[fig:3:exp30:a]{\fig\ref{fig:3:exp30:a}}&\checkmark \\
    (2)  &1.45\% & 0.00 & 10,000& 0.71\%& 0.92& 38.96\%& 0.91 & 42.70\% & 0.93 &\hyperref[fig:3:exp30:b]{\fig\ref{fig:3:exp30:b}}&\checkmark \\
    (3)  &10.59\% & 0.00 & 200& 10.63\%& 0.99 & 85.27\%& 0.99 & 87.86\% & 0.96 &\hyperref[fig:3:exp14:a]{\fig\ref{fig:3:exp14:a}}&\checkmark \\
    (4)  &14.38\% & 0.00 & 1,000& 2.81\%& 0.99 & 93.47\%& 0.99 & 94.19\% & 0.85 &\hyperref[fig:3:exp15:a]{\fig\ref{fig:3:exp15:a}}&\checkmark \\
    (5)  &5.74\% & 0.00 & 4,000& 10.00\%& 0.96 & 62.87\%& 0.96 & 68.74\% & 0.91 &\hyperref[fig:3:exp10:a]{\fig\ref{fig:3:exp10:a}}&\checkmark \\
    (6)  &1.57\% & 0.00 & 4,000& 0.47\%& 0.96 & 72.51\%& 0.96 & 81.03\% & 0.93 &\hyperref[fig:3:exp9:a]{\fig\ref{fig:3:exp9:a}}&\checkmark \\
    (7)  &5.74\% & 0.00 & 4,000& 10.00\%& 0.96 & 62.87\%& 0.96 & 68.74\% & 0.55 &\hyperref[fig:3:exp12:a]{\fig\ref{fig:3:exp12:a}}&\checkmark \\
    (8)  &0.64\% & 0.00 & 4,000& 1.06\%& 0.96 & 28.36\%& 0.97 & 64.88\% & 0.51 &\hyperref[fig:3:exp11:a]{\fig\ref{fig:3:exp11:a}}&\checkmark \\
    (9)  &1.00\% & 0.00 & 1,000& 0.91\%& 0.99 & 47.27\%& 0.98 & 50.86\% & 0.79 &\hyperref[fig:3:exp17:a]{\fig\ref{fig:3:exp17:a}}&\checkmark \\
    (10)  &12.02\% & 0.00 & 1,000& 11.04\%& 0.98& 55.56\%& 0.97 & 57.90\% & 0.60 &\hyperref[fig:3:exp16:a]{\fig\ref{fig:3:exp16:a}}&\checkmark \\
    \hline
    \multicolumn{6}{l}{(1) Default scenario \& \cifar~\cite{cifar}dataset} & \multicolumn{6}{l}{(2) Default scenario \& \gtsrb~\cite{gtsrb}dataset}\\
    \multicolumn{6}{l}{(3) Default scenario \& CNN with only FC layers, LR 0.0001, 3 epochs}& \multicolumn{6}{l}{(4) Default scenario \& CNN with batch normalization layers}\\
    \multicolumn{6}{l}{(5) \cifar~\cite{cifar} on \resneteighteen~\cite{resnet}, LR 0.001, 10 epochs} & \multicolumn{6}{l}{(6) \gtsrb~\cite{gtsrb} on \resneteighteen~\cite{resnet}, LR 0.001, ten epochs}\\
    \multicolumn{6}{l}{(7) \cifar~\cite{cifar} on \resnetthirtyfour~\cite{resnet}, LR 0.001, 10 epochs} & \multicolumn{6}{l}{(8) \cifarhundred~\cite{cifar} on \resneteighteen~\cite{resnet}, fine-tune on \cifar~\cite{cifar}, LR 0.001, 10 epochs}\\
    \multicolumn{6}{l}{(9) \cifarhundred~\cite{cifar} on VGG11~\cite{vgg_eleven}, LR 0.001, 200 epochs}& \multicolumn{6}{l}{(10) \cifarhundred~\cite{cifar} on ViT~\cite{vit}, LR 0.001, 5 fine-tuning epochs}\\
\end{tabular}
\vspace{-0.2cm}
\end{table*}

\subsection{Generalizability}
\label{sec:eval:appindependence}
Below, we explore different scenarios showing that \ourname can fulfill the generalizability requirement from \hyperref[sec:problem:objectives]{\sect\ref{sec:problem:objectives}}. Essentially we demonstrate the independence from datasets and model architectures. During these experiments, we use fine-tuning with \nicefrac{1}{10} of the original learning rate as the default model modification approach.

\begin{figure}[tb]
\centering
\begin{adjustbox}{max height=1.2cm}
\begin{subfigure}{0.18\linewidth}
    \includegraphics[width=\textwidth, cfbox=torstenbackground]{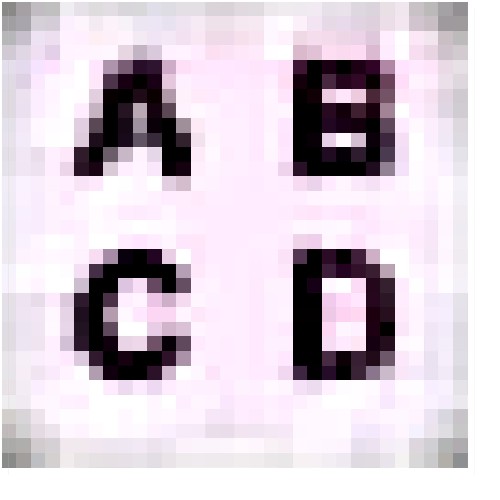} % 23_fc_mnist_3epochslowlr
    \vspace{-0.4cm}
\caption{}
    \label{fig:3:exp30:a}
\end{subfigure}
\hfill

\begin{subfigure}{0.05\linewidth}
    \includegraphics[width=\textwidth]{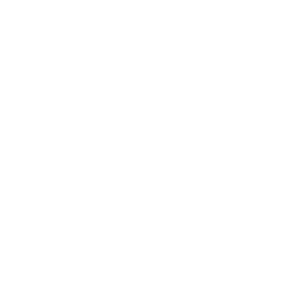} % 23_fc_mnist_3epochslowlr
    \vspace{-0.4cm}
\end{subfigure}
\hfill

\begin{subfigure}{0.18\linewidth}
    \includegraphics[width=\textwidth, cfbox=torstenbackground]{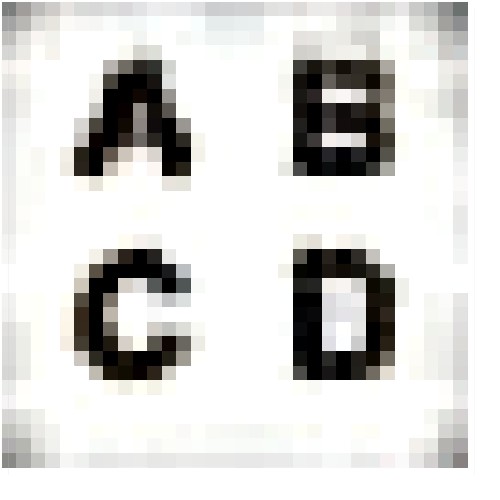} % 24_cnn_mnist_5epochs
    \vspace{-0.4cm}
\caption{}
    \label{fig:3:exp30:b}
\end{subfigure}
\hfill

\begin{subfigure}{0.05\linewidth}
    \includegraphics[width=\textwidth]{images/white.pdf} % 23_fc_mnist_3epochslowlr
    \vspace{-0.4cm}
\end{subfigure}
\hfill

\begin{subfigure}{0.18\linewidth}
    \includegraphics[width=\textwidth, cfbox=torstenbackground]{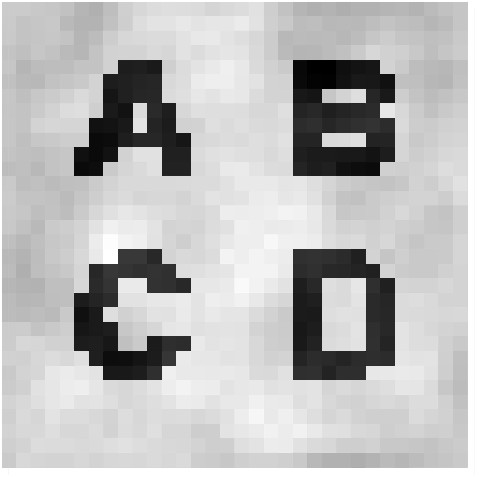} % 25_cnn_bn_mnist_5epochs
    \vspace{-0.4cm}
\caption{}
    \label{fig:3:exp14:a}
\end{subfigure}
\hfill

\begin{subfigure}{0.05\linewidth}
    \includegraphics[width=\textwidth]{images/white.pdf} % 23_fc_mnist_3epochslowlr
    \vspace{-0.4cm}
\end{subfigure}
\hfill

\begin{subfigure}{0.18\linewidth}
    \includegraphics[width=\textwidth, cfbox=torstenbackground]{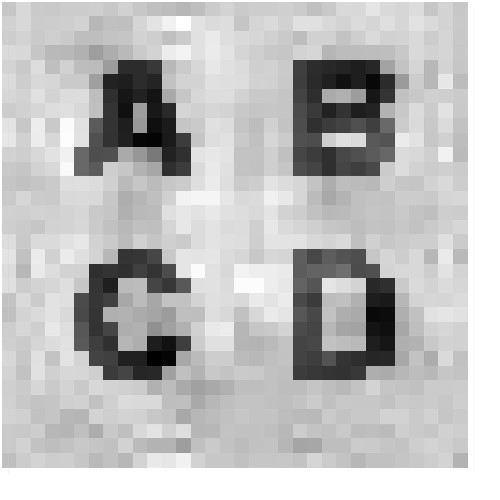} % 25_cnn_bn_mnist
    \vspace{-0.4cm}
\caption{}
    \label{fig:3:exp15:a}
\end{subfigure}
\hfill

\begin{subfigure}{0.05\linewidth}
    \includegraphics[width=\textwidth]{images/white.pdf} % 23_fc_mnist_3epochslowlr
    \vspace{-0.4cm}
\end{subfigure}
\hfill

\begin{subfigure}{0.18\linewidth}
    \includegraphics[width=\textwidth, cfbox=torstenbackground]{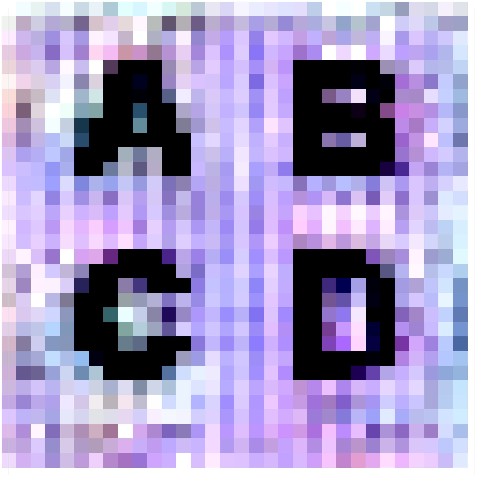} % 23_fc_mnist_3epochslowlr
    \vspace{-0.4cm}
\caption{}
    \label{fig:3:exp10:a}
\end{subfigure}
\end{adjustbox}
\begin{adjustbox}{max height=1.2cm}
\hfill
\begin{subfigure}{0.18\linewidth}
    \includegraphics[width=\textwidth, cfbox=torstenbackground]{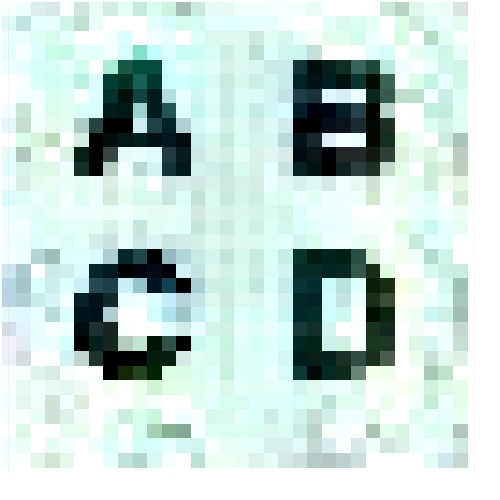} % 27_resnet34
    \vspace{-0.4cm}
\caption{}
    \label{fig:3:exp9:a}
\end{subfigure}
\hfill

\begin{subfigure}{0.05\linewidth}
    \includegraphics[width=\textwidth]{images/white.pdf} % 23_fc_mnist_3epochslowlr
    \vspace{-0.4cm}
\end{subfigure}
\hfill

\begin{subfigure}{0.18\linewidth}
    \includegraphics[width=\textwidth, cfbox=torstenbackground]{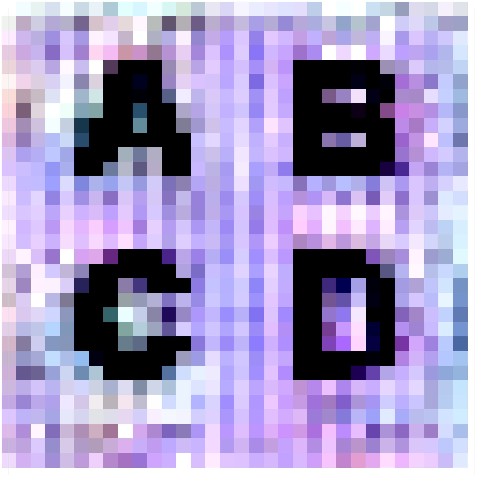} % 25_cnn_bn_mnist
    \vspace{-0.4cm}
\caption{}
    \label{fig:3:exp12:a}
\end{subfigure}
\hfill

\begin{subfigure}{0.05\linewidth}
    \includegraphics[width=\textwidth]{images/white.pdf} % 23_fc_mnist_3epochslowlr
    \vspace{-0.4cm}
\end{subfigure}
\hfill

\begin{subfigure}{0.18\linewidth}
    \includegraphics[width=\textwidth, cfbox=torstenbackground]{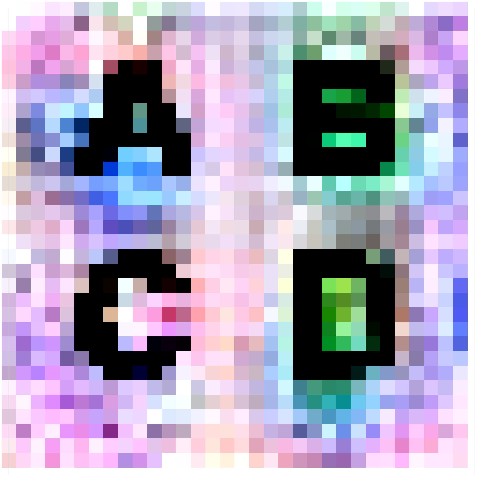} % 27_resnet34
    \vspace{-0.4cm}
\caption{}
    \label{fig:3:exp11:a}
\end{subfigure}
\hfill

\begin{subfigure}{0.05\linewidth}
    \includegraphics[width=\textwidth]{images/white.pdf} % 23_fc_mnist_3epochslowlr
    \vspace{-0.4cm}
\end{subfigure}
\hfill

\begin{subfigure}{0.18\linewidth}
    \includegraphics[width=\textwidth, cfbox=torstenbackground]{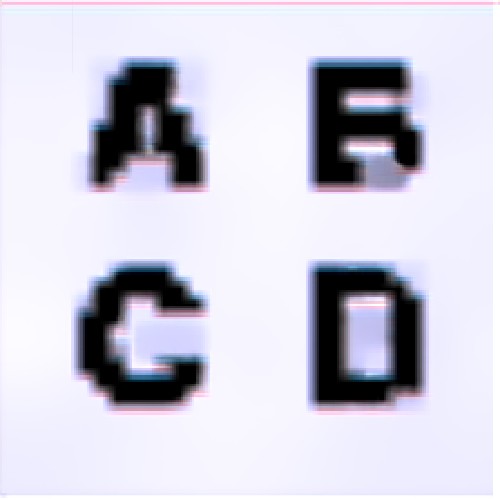}% {images/exp17_post_modification_lock_0.png} % 27_resnet34
    %\vspace{-0.2cm}
\caption{}
    \label{fig:3:exp17:a}
\end{subfigure}
\hfill

\begin{subfigure}{0.05\linewidth}
    \includegraphics[width=\textwidth]{images/white.pdf} % 23_fc_mnist_3epochslowlr
    \vspace{-0.4cm}
\end{subfigure}
\hfill

\begin{subfigure}{0.18\linewidth}
    \includegraphics[width=\textwidth, cfbox=torstenbackground]{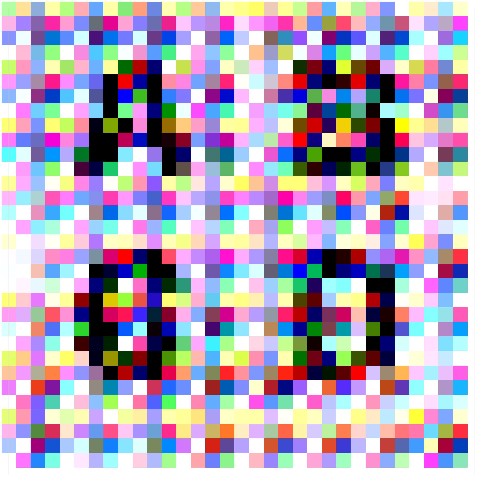} % 27_resnet34
    \vspace{-0.4cm}
\caption{}
    \label{fig:3:exp16:a}
\end{subfigure}

\end{adjustbox}
\caption{Visualization of extracted watermarks for the experiments listed in \hyperref[tab:datasetmodel]{\tab\ref{tab:datasetmodel}}.}
\vspace{-0.4cm}
\label{fig:3}
\end{figure}

\vspace{0.1cm}
\noindent \textbf{Dataset.} First, we changed the dataset to \cifar~\cite{cifar}, essentially changing the input layer to match the three color channels of the \cifar input samples. Then, we conducted the same experiment for \gtsrb~\cite{gtsrb}, which has 43 label classes. The results reported in (1) and (2) in \hyperref[tab:datasetmodel]{\tab\ref{tab:datasetmodel}} show, that the watermark was successfully embedded and survived fine-tuning yielding \hyperref[fig:3:exp30:a]{\fig\ref{fig:3:exp30:a}} and \hyperref[fig:3:exp30:b]{\fig\ref{fig:3:exp30:b}}. As the images clearly show the expected letters, we observe the dataset independence of \ourname.

\vspace{0.1cm}
\noindent \textbf{Small Model Architectures.} Next, to show that \ourname is also applicable to very simple model architectures, we trained \mnist~\cite{mnist} on a CNN with only three fully connected layers each of size 1024 and report the result in (3) in \hyperref[tab:datasetmodel]{\tab\ref{tab:datasetmodel}}. Further, to show \ournameGen independence of added batch normalization layers when embedding the watermark, we enhanced our default setting by adding such a layer after each convolutional layer  and report the results in (4) in \hyperref[tab:datasetmodel]{\tab\ref{tab:datasetmodel}}. The extracted images \hyperref[fig:3:exp14:a]{\fig\ref{fig:3:exp14:a}} and \hyperref[fig:3:exp15:a]{\fig\ref{fig:3:exp15:a}}, as well as the high SSIM values above 0.85 after raining and fine-tuning reported in \hyperref[tab:datasetmodel]{\tab\ref{tab:datasetmodel}} confirm \ournameGen good performance for small models. However, even if the images yield clear watermark evidence, the presence of batch normalization layers seems to diminish the robustness of the watermark resulting in a lower SSIM of 0.85 after fine-tuning. This effect might be caused by the circumstance, that for batch normalization layers, the mean and variance of the input data are unknown during transposed training and fixated to $E(x) = 0$ and $Var(x) = 1$, essentially causing information loss.

\vspace{0.1cm}
\noindent \textbf{Medium-Size Model Architectures.} To address bigger model architectures, we evaluate \cifar~\cite{cifar} and \gtsrb~\cite{gtsrb} on a \resneteighteen~\cite{resnet} model trained for ten epochs\footnote{As \resneteighteen~\cite{resnet} contains skip connections, we first trained the model for three epochs, to get valid values for fixating the skip connections, as discussed in \hyperref[sec:approach:transpose]{\sect\ref{sec:approach:transpose}}.}, whereas both datasets yielded similar results reported in (5) and (6) in \hyperref[tab:datasetmodel]{\tab\ref{tab:datasetmodel}}. As retractable in \hyperref[fig:3:exp10:a]{\fig\ref{fig:3:exp10:a}} and \hyperref[fig:3:exp9:a]{\fig\ref{fig:3:exp9:a}}, the watermark is still clearly visible after ten epochs.

Further, we evaluate \cifar~\cite{cifar} on \resnetthirtyfour~\cite{resnet} and use the same setup as in the \resneteighteen experiment. The results in (7) in \hyperref[tab:datasetmodel]{\tab\ref{tab:datasetmodel}} yield a bigger drop in SSIM to 0.55 after fine-tuning, probably introduced by the size of the model and the amount of transposed convolution layers, which introduce uncertainty due to their upsampling nature. Nevertheless, the resulting image \hyperref[fig:3:exp12:a]{\fig\ref{fig:3:exp12:a}} leaves no doubt that the watermark is still strongly embedded.

Inspired by~\cite{uchida2017embedding,adi2018turning}, we also evaluate cross-model fine-tuning scenarios. We used our medium-size model setups and trained \cifarhundred~\cite{cifar} on \resneteighteen~\cite{resnet} but fine-tuned the model on \cifar~\cite{cifar}, which necessitates changing the last layer due to ten instead of 100 output classes. As the watermark is embedded based on \cifarhundred, the last layer needs to be replaced with the original layer during watermark extraction, thus rendering this layer as part of the key. We report positive results with 0.51 SSIM in (8) in \hyperref[tab:datasetmodel]{\tab\ref{tab:datasetmodel}}. The watermark extraction yielded \hyperref[fig:3:exp11:a]{\fig\ref{fig:3:exp11:a}}, which clearly contains the watermark. This experiment demonstrates that the watermark is embedded within all network parameters and not only resides within the last layer, contributing to the robustness requirement from  \hyperref[sec:problem:objectives]{\sect\ref{sec:problem:objectives}}.

\vspace{0.1cm}
\noindent \textbf{Large Model Architectures.} To address even larger model architectures, we trained \cifarhundred~\cite{cifar} on VGG11~\cite{vgg_eleven} for 200 epochs and 100 epochs of fine-tuning. As the results in (9) in \hyperref[tab:datasetmodel]{\tab\ref{tab:datasetmodel}} show, \ourname could successfully embed a robust watermark even in a large model architecture. 

To evaluate \ourname on transformer blocks, we trained \cifar~\cite{cifar} on a Vision Transformer~\cite{vit}. The extracted watermark after fine-tuning is visualized in \hyperref[fig:3:exp16:a]{\fig\ref{fig:3:exp16:a}}. It has a colorful background, but the watermark text is clearly visible, indicating that \ourname can also handle such architectures.

\vspace{0.1cm}
\noindent Summarized, we showed, that \ourname is independent of the model architecture and, combined with the dataset independence, applicable in arbitrary application scenarios.

\begin{figure}[tb]
\centering
\begin{subfigure}{0.13\linewidth}
    \includegraphics[width=\textwidth, cfbox=torstenbackground]{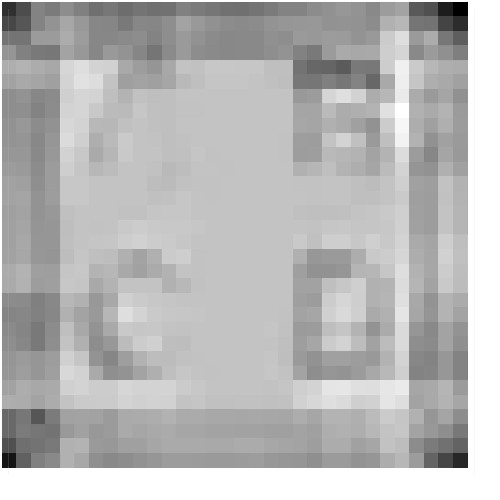} % 23_fc_mnist_3epochslowlr
\end{subfigure}
\hfill
\begin{subfigure}{0.13\linewidth}
    \includegraphics[width=\textwidth, cfbox=torstenbackground]{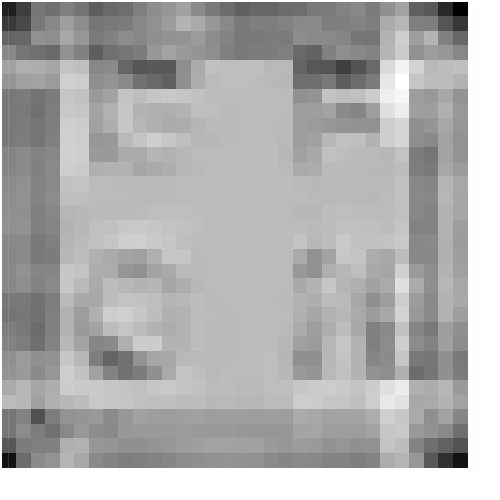} % 24_cnn_mnist_5epochs
\end{subfigure}
\hfill
\begin{subfigure}{0.13\linewidth}
    \includegraphics[width=\textwidth, cfbox=torstenbackground]{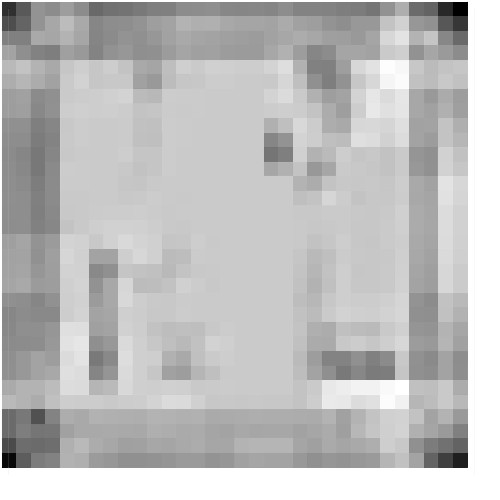} % 25_cnn_bn_mnist_5epochs
\end{subfigure}
\hfill
\begin{subfigure}{0.13\linewidth}
    \includegraphics[width=\textwidth, cfbox=torstenbackground]{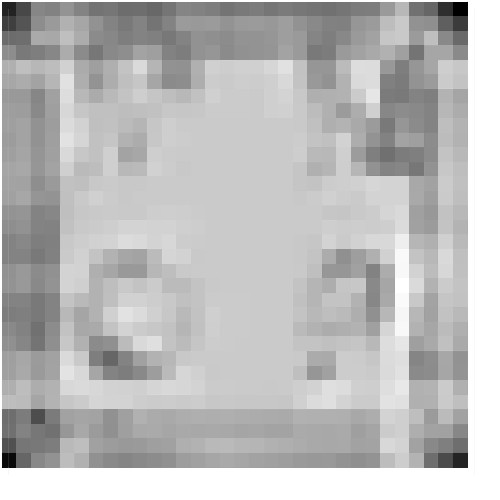} % 25_cnn_bn_mnist
\end{subfigure}
\hfill
\begin{subfigure}{0.13\linewidth}
    \includegraphics[width=\textwidth, cfbox=torstenbackground]{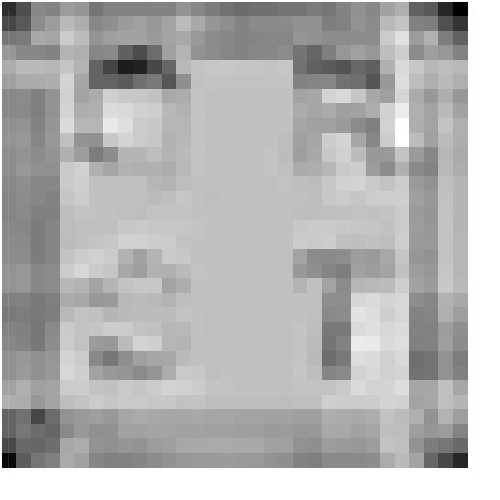} % 25_cnn_bn_mnist
\end{subfigure}
\hfill
\begin{subfigure}{0.13\linewidth}
    \includegraphics[width=\textwidth, cfbox=torstenbackground]{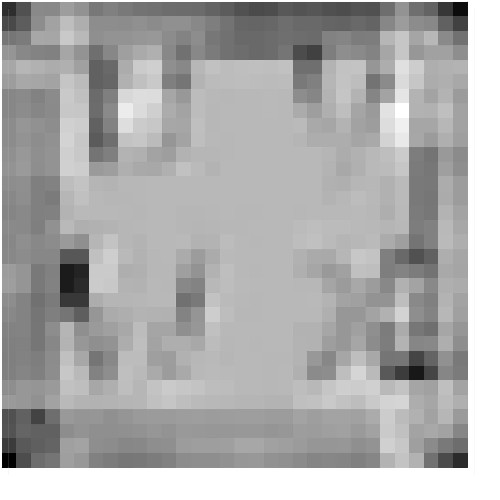} % 27_resnet34
\end{subfigure}
\hfill
\begin{subfigure}{0.13\linewidth}
    \includegraphics[width=\textwidth, cfbox=torstenbackground]{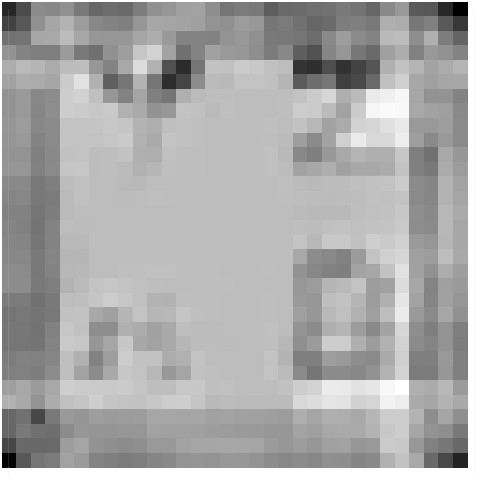} % 27_resnet34
\end{subfigure}

\begin{subfigure}{0.13\linewidth}
    \includegraphics[width=\textwidth, cfbox=torstenbackground]{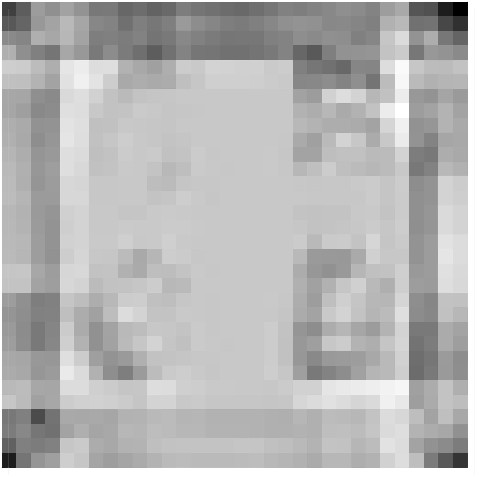} % 23_fc_mnist_3epochslowlr
\end{subfigure}
\hfill
\begin{subfigure}{0.13\linewidth}
    \includegraphics[width=\textwidth, cfbox=torstenbackground]{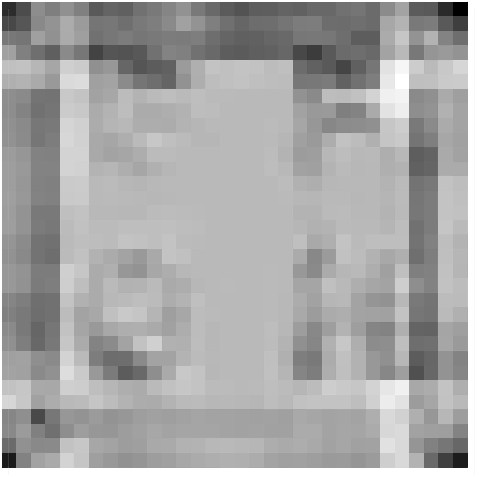} % 24_cnn_mnist_5epochs
\end{subfigure}
\hfill
\begin{subfigure}{0.13\linewidth}
    \includegraphics[width=\textwidth, cfbox=torstenbackground]{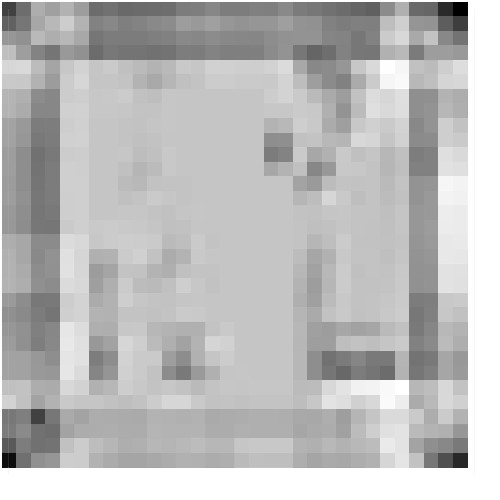} % 25_cnn_bn_mnist_5epochs
\end{subfigure}
\hfill
\begin{subfigure}{0.13\linewidth}
    \includegraphics[width=\textwidth, cfbox=torstenbackground]{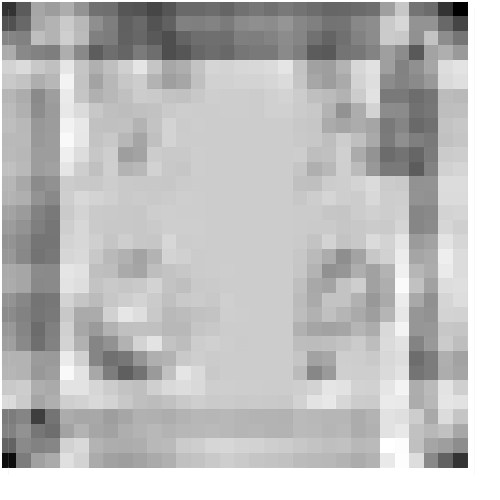} % 25_cnn_bn_mnist
\end{subfigure}
\hfill
\begin{subfigure}{0.13\linewidth}
    \includegraphics[width=\textwidth, cfbox=torstenbackground]{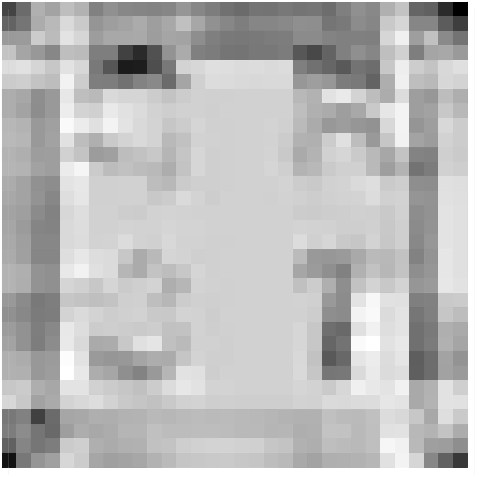} % 25_cnn_bn_mnist
\end{subfigure}
\hfill
\begin{subfigure}{0.13\linewidth}
    \includegraphics[width=\textwidth, cfbox=torstenbackground]{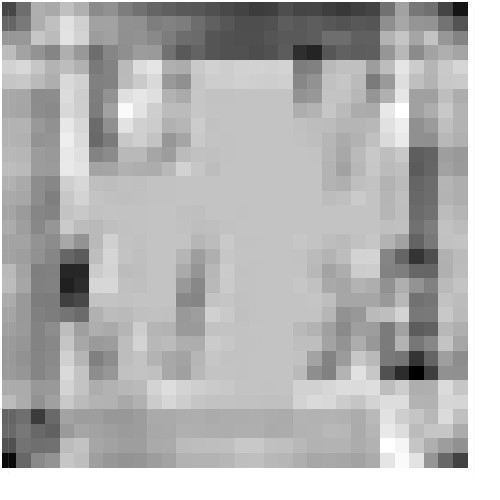} % 27_resnet34
\end{subfigure}
\hfill
\begin{subfigure}{0.13\linewidth}
    \includegraphics[width=\textwidth, cfbox=torstenbackground]{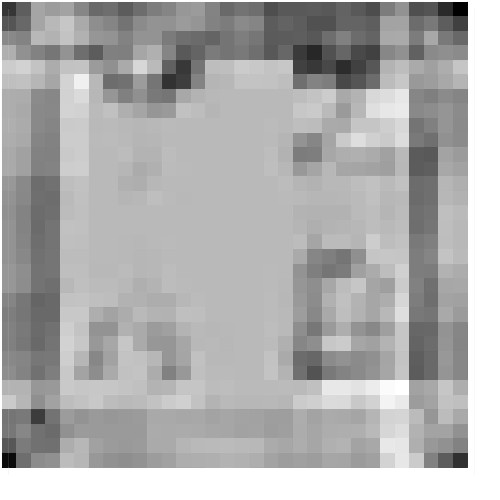} % 27_resnet34
\end{subfigure}

\begin{subfigure}{0.13\linewidth}
    \includegraphics[width=\textwidth, cfbox=torstenbackground]{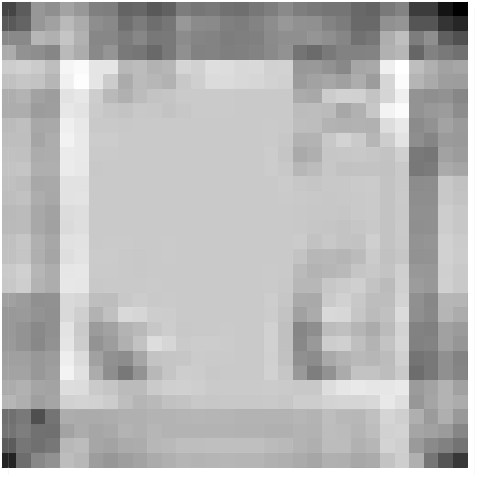} % 23_fc_mnist_3epochslowlr
\end{subfigure}
\hfill
\begin{subfigure}{0.13\linewidth}
    \includegraphics[width=\textwidth, cfbox=torstenbackground]{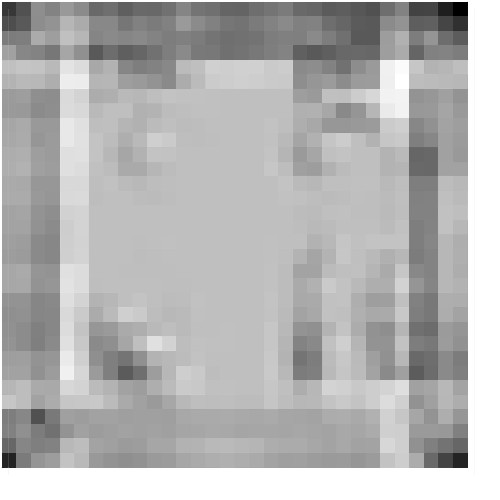} % 24_cnn_mnist_5epochs
\end{subfigure}
\hfill
\begin{subfigure}{0.13\linewidth}
    \includegraphics[width=\textwidth, cfbox=torstenbackground]{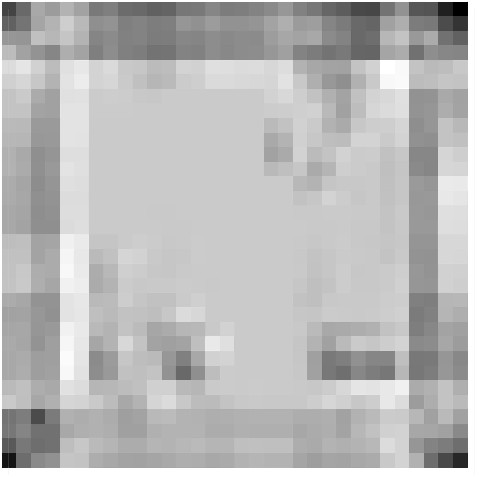} % 25_cnn_bn_mnist_5epochs
\end{subfigure}
\hfill
\begin{subfigure}{0.13\linewidth}
    \includegraphics[width=\textwidth, cfbox=torstenbackground]{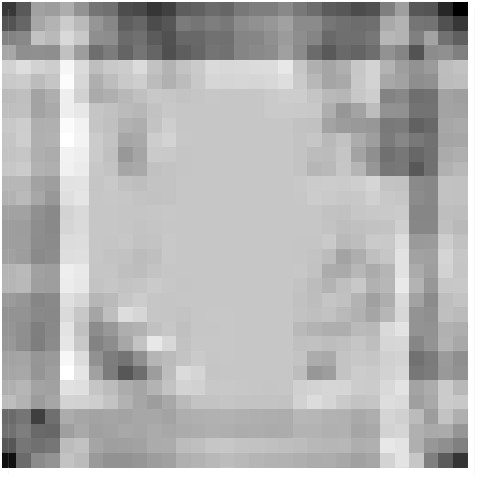} % 25_cnn_bn_mnist
\end{subfigure}
\hfill
\begin{subfigure}{0.13\linewidth}
    \includegraphics[width=\textwidth, cfbox=torstenbackground]{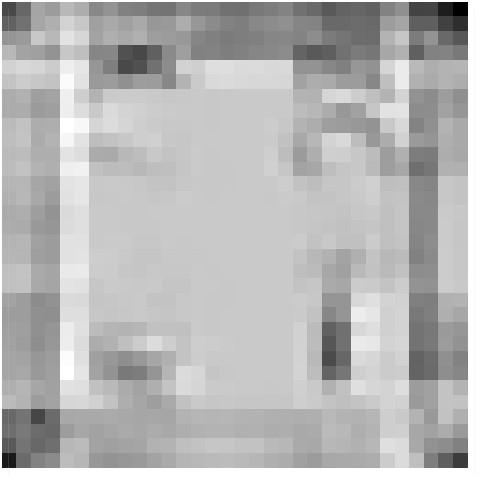} % 25_cnn_bn_mnist
\end{subfigure}
\hfill
\begin{subfigure}{0.13\linewidth}
    \includegraphics[width=\textwidth, cfbox=torstenbackground]{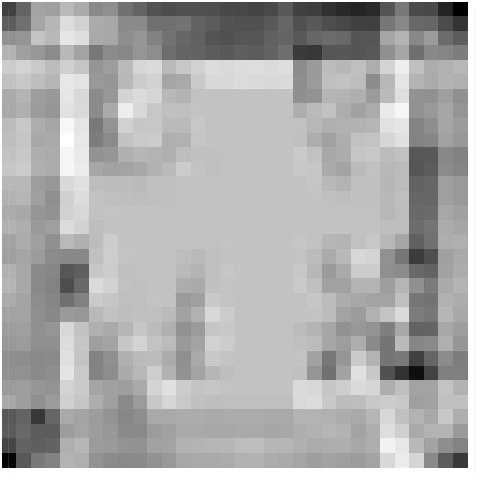} % 27_resnet34
\end{subfigure}
\hfill
\begin{subfigure}{0.13\linewidth}
    \includegraphics[width=\textwidth, cfbox=torstenbackground]{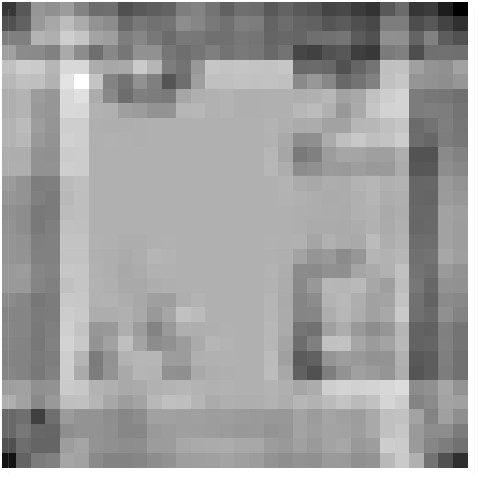} % 27_resnet34
\end{subfigure}
\caption{Extracted watermark images after five, seven, and eleven adversarial hardening steps during watermark removal. The corresponding main task accuracies to the lines are 82.89\%, 78.60\%, and 74.30\% originating from 89.10\% with SSIMs of 0.13, 0.11, and 0.01.}
\vspace{-0.5cm}
\label{fig:5}
\end{figure}

\vspace{-0.4cm}
\subsection{Adaptive Adversary}
\label{eval:adaptive}
\noindent \textbf{Watermark Erasure.} An adversary with knowledge of \ournameGen functionality could try to erase an embedded watermark as defined in \hyperref[sec:problem:threatmodel]{\sect\ref{sec:problem:threatmodel}}. Thereby, a random watermark key in the predefined value range could be used in combination with a secret image consisting of random noise. An even stronger adversary with the knowledge of embedded watermark keys (which exceeds usual assumptions as in \hyperref[sec:problem:threatmodel]{\sect\ref{sec:problem:threatmodel}} could also use such an image.
As both scenarios yield the same effects, we depict the results for a stronger adversary (using an already embedded watermark key) in the main body of the paper and append the random key experiment results in \hyperref[app:additionalexperiments]{\appSect\ref{app:additionalexperiments}}. 

We use the same mechanism as in step 1 in \hyperref[fig:overview]{\fig\ref{fig:overview}} to embed the adversarial key-secret pair,  with the intention of erasing the existing secret for the respective watermark key and potentially for other existing keys that are part of the watermark. Our experiment revealed, that the erasure of the watermark involves a significant loss in main task accuracy as both tasks are entangled within the model parameters by design. During watermark erasure attempts, the adversary sacrifices usually more than 10\% accuracy in our setup compared to the initial 89.10\%, whereas some other existing works consider 3.5\% accuracy drop as acceptable in related works~\cite{lemerrer2019frontierstitching}. In \hyperref[fig:5]{\fig\ref{fig:5}}, we show seven out of eleven watermark images\footnote{We show only seven images due to space limitations in the paper.} after five, seven, and eleven adversarial hardening steps with respective remaining accuracy values of 82.89\%, 78.60\%, and 74.30\%. Even the third row with an accuracy drop of 14.8\% shows clear evidence, as one should keep in mind, that an unwatermarked model yields an image similar to \hyperref[fig:1:exp2:a]{\fig\ref{fig:1:exp2:a}}. In the first and second rows of \hyperref[fig:5]{\fig\ref{fig:5}}, the watermark existence can still be attested by a human observer even though the SSIMs are low with 0.13, 0.11, and 0.01. This shows that the definition of rigid thresholds is challenging as in the case of SSIMs such low values could also stem from content-wise unrelated images. Therefore, a threshold needs to be set higher to avoid false positives. Hence, \ourname improves the decision-making in such situations, essentially increasing the security for the model owner while fulfilling the understandability requirement from \hyperref[sec:problem:objectives]{\sect\ref{sec:problem:objectives}}. We got similar results when using a completely black image as a key and when using an image that an unwatermarked model yields for the adversarial key after a transposed inference, which is reported in \hyperref[app:additionalexperiments]{\appSect\ref{app:additionalexperiments}}. 

\begin{figure}[tb]
\centering
\begin{subfigure}{\linewidth}
    \includegraphics[width=\textwidth]{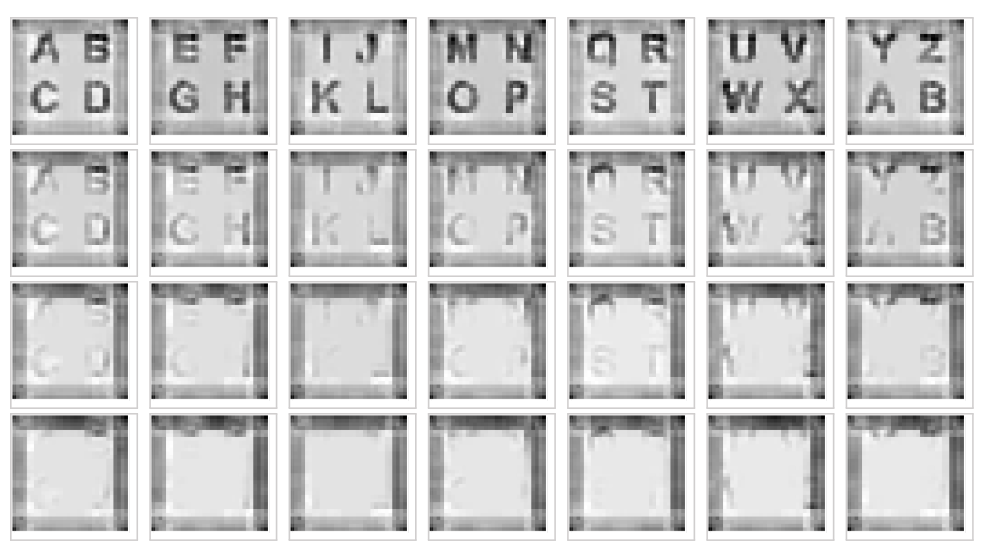} % 23_fc_mnist_3epochslowlr
\end{subfigure}
\caption{Watermarks after watermark overwriting with eleven keys. The lines correspond to the first four hardening steps with accuracy drops from 89.10\% to 83.89\%, 70.38\%, 56.39\%, and 46.52\%, and SSIMs of 0.52, 0.17, 0.05, 0.01.}
\vspace{-0.5cm}
\label{fig:elevenrandom}
\end{figure}

Further, we can report, that we observed the same effect when increasing the number of keys embedded by the adversary from one to eleven. We experimented with eleven already embedded or eleven random keys combined with secret images containing random noise, only black pixels, and yielded images from an unwatermarked model like \hyperref[fig:1:exp2:a]{\fig\ref{fig:1:exp2:a}}.\footnote{Due to space limitations in the paper we only report one scenario, as all scenarios yield similar results.} For example, for the latter, we measured accuracy drops from 89.10\% to 83.89\%, 70.38\%, 56.39\%, and 46.52\% after one to four hardening steps resulting in the four lines of \hyperref[fig:elevenrandom]{\fig\ref{fig:elevenrandom}}, where the watermark is still completely identifiable in the second line, and partially even in the third line. To achieve a removal degree as showcased in the fourth line, the adversary would need to sacrifice 42.58\% in main task accuracy.

\begin{figure}[tb]
\centering
\begin{subfigure}{\linewidth}
    \includegraphics[width=\textwidth]{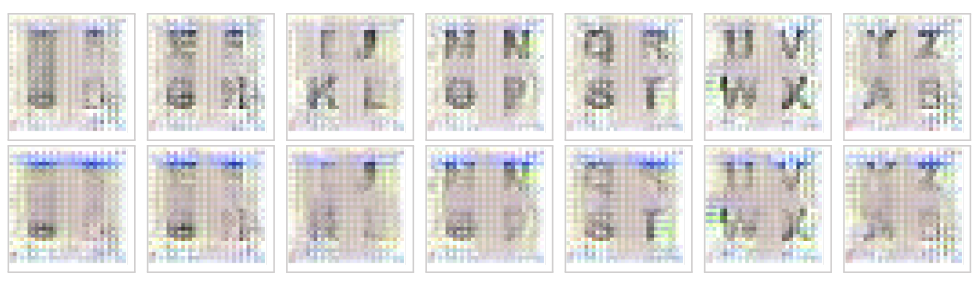} % 24_cnn_mnist_5epochs
\end{subfigure}
\caption{Extracted watermarks after a watermark erasure attack on \resneteighteen~\cite{resnet} trained on \cifar~\cite{cifar}. The lines show watermarks after eleven (line one) and 22 (line two) hardening steps corresponding to 10.12\% and 18.62\% accuracy drop and SSIMs of 0.38 and 0.24.}
\vspace{-0.5cm}
\label{fig:cifarextract}
\end{figure}

To validate that our findings are not intrinsic to our specific application scenario, we also conducted the experiment with an already embedded watermark key and an inferred image from an unwatermarked model similar to \hyperref[fig:1:exp2:a]{\fig\ref{fig:1:exp2:a}} for \cifar~\cite{cifar} trained on \resneteighteen~\cite{resnet}. Thereby, we could observe the same effect of steadily decreasing main task accuracy with increasing watermark elutriation. We report the watermarks after eleven and 22 hardening steps with accuracy drops of 10.12\% and 18.62\% percent in \hyperref[fig:cifarextract]{\fig\ref{fig:cifarextract}}, showing that the watermark is still partially embedded while the adversary sacrificed a significant amount of main task accuracy.

\begin{figure}[tb]
\centering

\begin{subfigure}{\linewidth}
    \includegraphics[width=\textwidth]{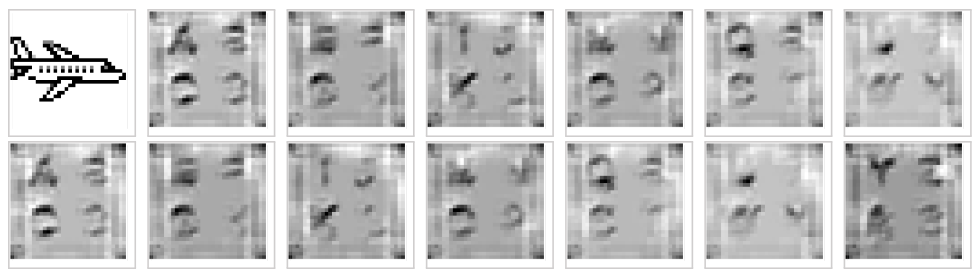} % 23_fc_mnist_3epochslowlr
\end{subfigure}

\caption{An adversarial watermark secret (first image) and extracted watermarks after six (line one) and seven (line two) hardening steps corresponding to 79.25\% and 78.25\% remaining main task accuracy originating from 89.10\% and SSIMs of 0.24 and 0.23.}
\vspace{-0.5cm}
\label{fig:airplane}
\end{figure}

\vspace{0.1cm}
\noindent \textbf{Watermark Overwriting.} Besides removing the watermark, an adversary could also try to overwrite the watermark with a second watermark. When embedding one new watermark image, in our case an airplane icon as visible in the first image in the first line of \hyperref[fig:airplane]{\fig\ref{fig:airplane}} with a random key, the adversary provoked an accuracy drop from 89.10\% to 79.25\% and 78.25\% after six and seven hardening steps. While evidence for the original watermark is still visible after six steps, it starts to be vague after seven steps, as can be seen in the first and second lines in \hyperref[fig:airplane]{\fig\ref{fig:airplane}}, respectively.

\vspace{0.1cm}
\noindent Summarized, we can conclude, that \ourname is robust against adaptive adversaries with the knowledge of \ournameGen functionality, that try to remove or overwrite the watermark or embed an additional watermark, even if the adversary is equipped with knowledge about the keys. If the adversary erases the watermark, he sacrifices a minimum of 10\% of accuracy, whereas usually around 3.5\% accuracy drop is considered as acceptable~\cite{lemerrer2019frontierstitching}.

\subsection{Capacity}
\label{eval:capacity}

To evaluate the capacity and generalizability to arbitrary watermark locks, we generated random bit strings and converted them into images using dot code ~\cite{gils1987dotcode}. The image is initially divided into square patches that correspond to the number of bits in the string. Based on the bit value, the patch is colored black for zero or white for one. An example of this is shown in \hyperref[fig:watermark_capacity_example]{\appFig\ref{fig:watermark_capacity_example}}. Multiple watermarks were injected into the model, with the same capacity and randomly generated keys. Likewise, the extracted watermark was again divided into patches to extract the bit sequence. Values less than 0.53 in each channel were set to 0, while others were set to 1. To determine the bit of the patch, black and white pixels were counted and the majority was used as a decoding result. To evaluate the outcome, the Binary Error Rate (BER) was utilized, which is the number of incorrect bits divided by the total number of bits. We first generated images with 36-bit capacity and injected those into the CNN model. The evaluation has shown that the BER is consistently low at roughly 3.43\% and 2.96\% after fine-tuning. For larger bit lengths of 100 bits per image we injected the MiT license text \footnote{MiT License available at \url{https://opensource.org/license/mit/}.} (8,544 bits) into a \resneteighteen~\cite{resnet} model. We compared the results for applied (7,4) Hamming code~\cite{hamming1950hammingcode} error correction (14,952 bits) and without. \ourname achieves average BERs after training of as little as approximately 5.92\% and 3.62\% with and without error correction, respectively. After fine-tuning the BER increased to 6.46\% without error correction and \errorrate with error correction. \ourname is capable of injecting large payloads such as licenses or images and, as the limit of keys is not yet exhausted, even larger files could be injected.

\vspace{-0.4cm}
\subsection{Runtime}
Regarding the runtime of \ourname, we measured the individual steps, namely hardening, constraint training, and watermark testing for our default scenario with eleven random keys. We report averages over ten experimental runs. Hardening with 10,000 steps took 53.48 seconds and is a one-time effort. The evaluation only takes 0.02 seconds. The training time increased from 67.62 to 94.39 seconds for the five epochs, introducing a one-time overhead of 39.58\%, which is expected, as an additional loss needs to be computed and a second parameter optimization step is executed during training. Therefore, we consider the efficiency requirement from \hyperref[sec:problem:objectives]{\sect\ref{sec:problem:objectives}} as fulfilled.

\noindent Summarized, we can show that \ourname is a robust watermarking mechanism that withstands adversarial attempts to remove the watermark and barely influences the main task performance of the \dnn while providing high watermark robustness. Most importantly, no non-intuitive algorithm based on a riding threshold needs to be applied to evaluate the watermark.

\section{Discussion}
\label{sec:discussion}
In this section, we discuss potential approaches for automated ownership claim validation (\sect\ref{sec:discussion:automation}), as well as elaborate on constraint optimization methods that could be alternatively employed (\sect\ref{sec:discussion:optimization}).

\subsubsection{Ownership Claim Automation} 
\label{sec:discussion:automation}
Ideally, a human evaluator should determine the ownership claim and the presence of a watermark by comparing it to the ground truth. However, in situations where a large volume of images needs validation, automation becomes essential.
One potential automated approach involves employing similarity metrics to assess whether a watermark matches the ground truth watermark. However, relying solely on the Structural Similarity Index (SSIM) metric may not be optimal. The evaluation demonstrates that the watermark remains visually detectable even at low SSIM values.

A more effective method could involve automating the decision-making process through machine learning (ML). This could be achieved by training an ML model to serve as a feature extractor, responsible for generating embeddings from both the watermark image and the ground truth watermark. By comparing these embeddings using a decision layer within the network, the system can make a final determination about the presence of a copyright infringement. This automated approach could ensure accurate and efficient validation, especially when dealing with a large number of images.

\subsubsection{Other Constraint Optimization Methods}
\label{sec:discussion:optimization}
As an alternative to the sequential optimization described in \hyperref[sec:approach:moo]{\sect\ref{sec:approach:moo}}, one could leverage a weighted sum method, which adds up the two losses while assigning corresponding weights, essentially introducing an additional hyperparameter. Usually, such weights indicate the importance of the tasks. However, this method necessitates extensive hyperparameter tuning if the loss values are at different scales. The loss values then need to be weighted such that both loss terms receive equal importance, since otherwise the smaller one is deemed already sufficiently optimized. 

Constraint optimization methods, e.g., Augmented Lagrangian optimization~\cite{ala} or the Alternating Direction Method of Multipliers~\cite{boyd2011distributed} are highly effective in enforcing hard constraints. However, it is important to note that our approach, \ourname, does not impose strict thresholds for valid 
SSIM values. Introducing these methods would necessitate adding thresholds and hyperparameters to \ourname, which would then need optimization. It's worth emphasizing that a rigid SSIM threshold isn't essential in our context; we aim for high SSIM values around 0.9, but the similarity between an extracted watermark and the secret can already be claimed for very low SSIMs. Implementing such constraint optimization methods would shift the focus toward optimizing the watermarking task to meet the defined threshold. This could potentially detract attention from the primary task and lead to suboptimal performance in both tasks. Moreover, after the constraint is met, there's a risk of diminishing the importance of the watermarking task, potentially hindering the achievement of superior quality.

However, in \mbox{real-world} applications where a hard threshold is desired, one might consider adapting the approach presented in~\cite{ala}, as it is a method capable of reliably enforcing inequality constraints.

\section{Related Work}
\label{sec:relatedwork}
Our approach, \ourname, stands out as the pioneer in the field of model watermarking by incorporating transposed model training, making it unique and unparalleled in comparison to existing methodologies. This novel technique sets it apart from related research, rendering direct comparisons challenging. The distinctive advantage of our method lies in its departure from conventional approaches that rely on fixed thresholds for decision-making during watermark verification. Instead, \ourname produces a discernible image with interpretable content. This output can be easily scrutinized by human evaluators, enabling intuitive decision-making when compared to the authentic ground truth image.

In the following sections, we present an overview of contemporary techniques in model watermarking and fingerprinting. Despite employing diverse methods, these existing approaches share common objectives with ours. Typically, watermarking methods seek to embed a distinctive signature into the model, ensuring its uniqueness to the model owner. On the other hand, fingerprinting methods are geared toward copyright protection, embedding a unique signature into the model that is specific to the authorized user. Additionally, we overview the related works that, while not strictly falling under watermarking or fingerprinting categories share relevance due to their underlying methodologies or the goals they aim to accomplish.

\vspace{0.1cm} 
\noindent\textbf{Watermarking.} Uchida~\etal~\cite{uchida2017embedding} proposed a white-box, multi-bit watermarking approach, which embeds the watermark into weights of convolutional layers by introducing an additional loss term, referred to as parameter regularization. However, the method relies on a rigid threshold for watermark verification and is not robust against watermark overwriting attacks. ST-DM~\etal~\cite{li2021spreadtransform} improves this approach in such scenarios by using modulation techniques. The capacity of the watermark was thereby shown for up to 8,400 bits but naturally is limited by the model architecture.

Frontier Stitching ~\cite{lemerrer2019frontierstitching} is a 1-bit, black-box methodology rooted in adversarial examples. 
This technique modifies the decision boundaries between classes for specific input samples positioned at the interface of two classes, serving as essential keys. However, identifying such samples is contingent upon the specific application context and proves to be a challenging task, making this approach less universally applicable and user-friendly. For verification, a hard threshold of prediction matches is used. 

Adi~\etal~\cite{adi2018turning} propose a multi-bit, black-box method, which embeds backdoors into the DNN that serve as a watermark. Within the paper, eight backdoors acting as watermarks were implanted, whereas randomly generated input samples were used to produce specific output classes. Similarly, Zhang~\etal~\cite{zhang2018protecting} and Zhang~\etal~\cite{zhang2020modelwatermarking} follow the same approach varying the backdoor types, whereas Li~\etal~\cite{zheng2019blindwatermark} produces an imperceptible backdoor trigger with a second generative model. A method that also leverages backdoor-like behavior is Guo~\etal~\cite{guo2018watermarking}, where a black-box watermarking technique is proposed specially crafted for embedded devices. The method trains the DNN to behave significantly differently on a portion of training samples, that are modified using a specific perturbation. Naturally, these approaches embed additional behavior besides the model's main task, providing the potential for side effects during normal inference. Further, the decision-making relies on a threshold for the number of occurred backdoor-based mispredictions. Contrary to these works, WILD~\cite{liu2021removing} is an approach that removes backdoor-based watermarks.

Tartaglione~\etal~\cite{tartaglione2021delving} propose a white-box, 1-bit watermark that fixates model parameters as watermarks and applies a modified loss function for increased robustness. As the approach fixates parameters, the watermark capacity is limited by the number of parameters in the model and hence depends on the model architecture.

DeepJudge~\cite{chen2022deepjudge} is a testing framework that can be used for copyright protection as a non-invasive alternative to watermarking techniques. The method compares how similar a DNN and a second suspected DNN under test behave based on six metrics and hard thresholds. The metrics are derived via inference of carefully chosen samples that are able to characterize the models. 

Wang~\etal~\cite{wang2020watermarkingbackpropagation} presents a white-box approach for incorporating a multi-bit watermark into weights by leveraging a second secret independent model for watermark embedding and verification. Similarly, RIGA~\cite{wang2021riga} is an algorithm that embeds a multi-bit, white-box watermark using adversarial training with two additional models. Thereby, the first model is responsible for embedding the watermark, while the second enhances the stealthiness. However, the verification of both approaches is based on the black-box functionality of additional models. Hence, the decision-making is not comprehensible, as the model's functionality does not follow an understandable algorithmic pattern, but delivers outputs utilizing optimized parameters tuned with regard to training data.

EWE~\cite{jia2021entangledwm} is a method based on a special loss function, which enforces the entanglement of the watermark and the main task (similar to \ourname), such that removing the watermark negatively affects the main task. However, the watermark is embedded in the forward path, thus leaving the possibility of unexpected side-effects during inference.

DeepSigns~\cite{rouhani2019deepsigns} proposes a white-box, multi-bit approach that embeds a watermark within the probability density function of activations in multiple DNN layers by fine-tuning the model parameters. When verifying the watermark, an algorithm uses the extracted activations to compute the watermark which is then evaluated against the ground truth utilizing bit error rate combined with a rigid threshold. Further, a black-box approach is suggested, that verifies model ownership by a hard threshold applied to the number of matches when comparing the prediction outputs of specific secret input-prediction pairs. 

\vspace{0.1cm} 
\noindent\textbf{Fingerprinting.} A consecutive work to DeepSigns~\cite{rouhani2019deepsigns} leveraging the same principle and inheriting the same shortcomings is DeepMarks~\cite{chen2019deepmarks}, but embeds information in the model weights instead of the activations and is designed as a fingerprinting approach. Based on this fingerprinting technique, DeepAttest~\cite{chen2019deepattest} offers hardware-level IP protection and usage control for DNNs. With the help of a Trusted Execution Environment, the fingerprint is validated to ensure that only validated DNNs are allowed to run on specific devices, a method that is also leveraged in DeepMark~\cite{xie2021deepmark}. 

IPGuard~\cite{xiaoyu2021ipguard} is a method that searches for adversarial examples that are close to the decision boundary of a DNN, leveraging that a model is characterized by its decision boundary. The method does not tamper with the training process at all, but can run after training even on legacy models and, hence, does not affect the model performance at all. However, humans who do not understand the decision boundaries of DNNs might have problems understanding the approach. Further, the ownership verification is based on a hard threshold, which is difficult to determine. The decision boundary is also leveraged by MetaFinger~\cite{yang2022metafinger}, a black-box fingerprinting method that identifies samples by meta-training that are close to the decision boundary, which can later be used to identify a specific model.

\vspace{0.1cm} 
\noindent\textbf{Orthogonal Works.} Further, there are some works, that are close but also orthogonal to \ourname. TamperNN~\cite{merrer2019tampernn} is a method designed to recognize if a model was tampered, e.g., fine-tuned, by analyzing inputs, that tend to change the prediction class easily. Chen~\etal~\cite{chen2022teacher} suggest a method to infer the origin of a student model in the domain of transfer learning by embedding a fingerprint in the teacher model that is passed on to the derived student model. Venugopal~\etal~\cite{venugopal2011watermarkingoutputs} propose a method to watermark the model output instead of the model itself by selecting a specific result out of the selection of possible results in a machine translation task. DAWN~\cite{Szyller2021dawn} is a technique used to prevent model extraction attacks by changing the prediction on the model inference API for a small set of samples to embed a watermark into models trained on these predictions. BOP~\cite{cortinas-lorenzo2020adam} modifies the Adam optimizer to prevent so-called heavily spiked weights during watermark embedding and, hence, increase covertness while simultaneously increasing the robustness. Wang~\etal~\cite{wang2019attacksonwm} demonstrated that statistical analysis of model weights can detect a watermark. Once identified, the watermark can be overwritten using the original embedding technique, effectively removing it.

Deconvolutions~\cite{zeiler2010deconvolutional,pixeltransposed} are the inspiration and basis for our work and are used to approximately reverse convolutional operations that are often leveraged in machine learning models. Such deconvolutions are used in scenarios that require up-sampling of feature maps, such as generative models~\cite{deconvgan,deconvgantwo} and encoder-decoder architectures~\cite{im2019dt,autoencoderone}, which generate images from an embedding. However, the deconvolutions mostly possess their own trainable weights independent of the convolutions. Weight sharing is used in Siamese Networks~\cite{koch2015siamese}, which provides the motivation to share the weights between convolutions and deconvolutions and, thus, for transposed training.

\section{Conclusion}
\label{sec:con}
Machine learning models can be considered as intellectual property of the model creator that needs to be protected from unauthorized use by third parties. DNN watermarking techniques offer a solution to this problem by embedding a secret watermark into the model parameters. Obviously, these watermarks must be robust against erasure attempts, while simultaneously a minimal effect on the model's main task is expected.

Existing watermarking approaches rely on rigid thresholds in the final decision-making process after the watermarking data is extracted from the model during watermark verification. Thereby, such a threshold can fail to detect remaining fractions of embedded watermarks that were attacked with an erasure attempt, even if a human observer would clearly identify the remaining watermark.

To address this problem we proposed \ourname, the first human-understandable and intuitive DNN watermarking approach that allows human decision-making directly on the extracted watermark data without relying on a threshold. We show that \ournameGen effect on the model's performance is negligible and that \ourname is independent from specific application scenarios. Further, \ourname withstands adversarial model manipulations and offers a capacity of \bitnumber bits with a low error rate of \errorrate.
%\textbf{13 page limit}

\vfill
\bibliographystyle{plain}
\bibliography{references}

\balance

\clearpage
\section*{Appendix}
In this section, we present additional experiment details, such as randomly generated key vectors and watermark images in Sections \hyperref[app:additionaldetails]{\sect\ref{app:additionaldetails}}. Moreover, the impact of using different optimizers and further watermark removal experiments are discussed in \hyperref[app:additionalexperiments]{\sect\ref{app:additionalexperiments}}.

\subsection{Additional Experimental Details}
\label{app:additionaldetails}

\vspace{0.1cm} \noindent 
\textbf{Random Key Vectors.} The key vectors are randomly chosen between -10 and 10. Thereby it is not imperative, that -10 and 10 need to be part of the vector. For example, the key vector from the single watermark experiments reported within this paper consists of the following values: -0.0748, 5.3644, -8.2304, -7.3593, -3.8515, 2.6815, -0.1981, 7.9288, -0.8874, 2.6461.

\begin{figure}[tb]
\centering
\begin{subfigure}{\linewidth}
    \includegraphics[width=\textwidth]{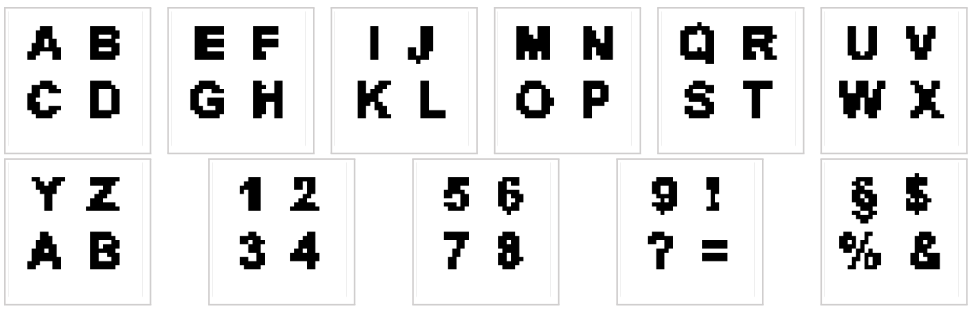} % 1_only_wm_final
\end{subfigure}
\caption{Visualization of the default eleven watermark secrets (images) used within the main part of this paper.}
\label{fig:secrets_default}
\end{figure}

\begin{figure}[tb]
\centering
\begin{subfigure}{\linewidth}
    \includegraphics[width=\textwidth]{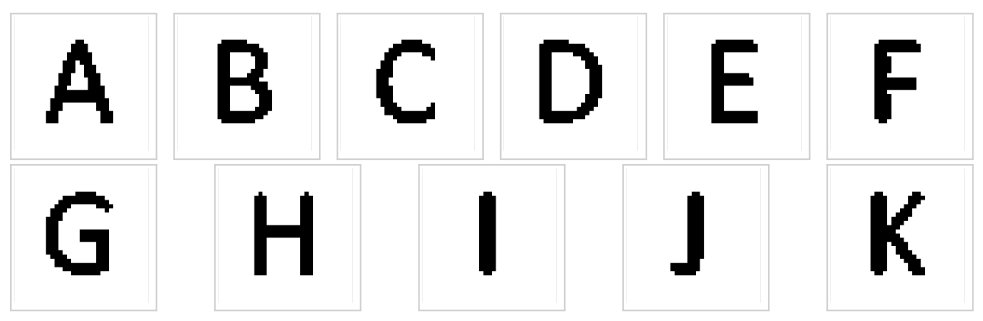} % 1_only_wm_final
\end{subfigure}
\caption{Visualization of eleven watermark secrets (images) used as an alternative to the default images in the main part of this paper. The images use one letter instead of four on each image.}
\label{fig:secrets_airplane}
\end{figure}

\begin{figure}[tb]
\centering
\begin{subfigure}{\linewidth}
    \includegraphics[width=\textwidth]{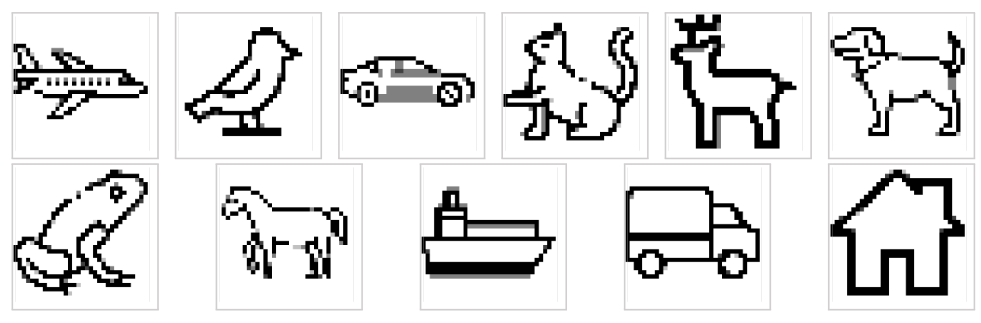} % 1_only_wm_final
\end{subfigure}
\caption{Visualization of eleven watermark secrets (images) used as an alternative to the default images in the main part of this paper. The images are inspired by the \cifar~\cite{cifar} classes.}
\label{fig:secrets_airplane_real}
\end{figure}

\vspace{0.1cm} \noindent 
\textbf{Watermark Secrets.} During our experiments, we use eleven distinct secrets. The secrets are images with black text on a white background, as visualized in \hyperref[fig:secrets_default]{\fig\ref{fig:secrets_default}}. Additionally, we conducted an experiment with different watermark images, which yielded the same experimental outcomes. Thereby, we used only one letter on each image (\cf~\hyperref[fig:secrets_airplane]{\fig\ref{fig:secrets_airplane}}) as well as the icons of the ten \cifar~\cite{cifar} label classes and added an extra icon as visualized in \hyperref[fig:secrets_airplane_real]{\fig\ref{fig:secrets_airplane_real}}.

\begin{figure}[tb]
\centering
\includegraphics[width=0.25\textwidth]{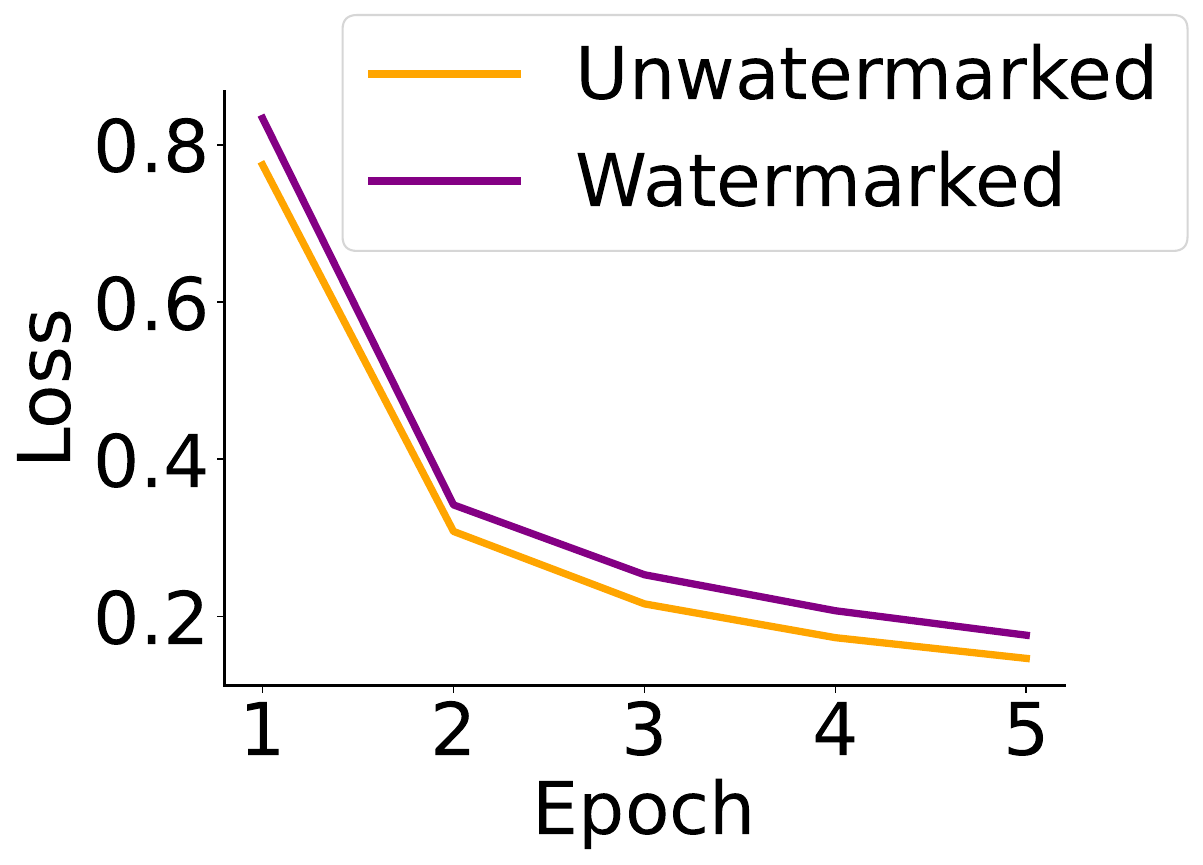}
\caption{Main task loss during training of an (a) unwatermarked and a (b) watermarked model.}
\label{fig:mainlosses}
\end{figure}

\vspace{0.1cm} \noindent 
\textbf{Fidelity.} To highlight the fidelity of \ourname we provide an additional plot of the main task loss during training of an unwatermarked and a watermarked model yielding minimal differences in \hyperref[fig:mainlosses]{\fig\ref{fig:mainlosses}}.

\begin{figure}
    \centering
    \includegraphics[width=0.1\textwidth, cfbox=torstenbackground]{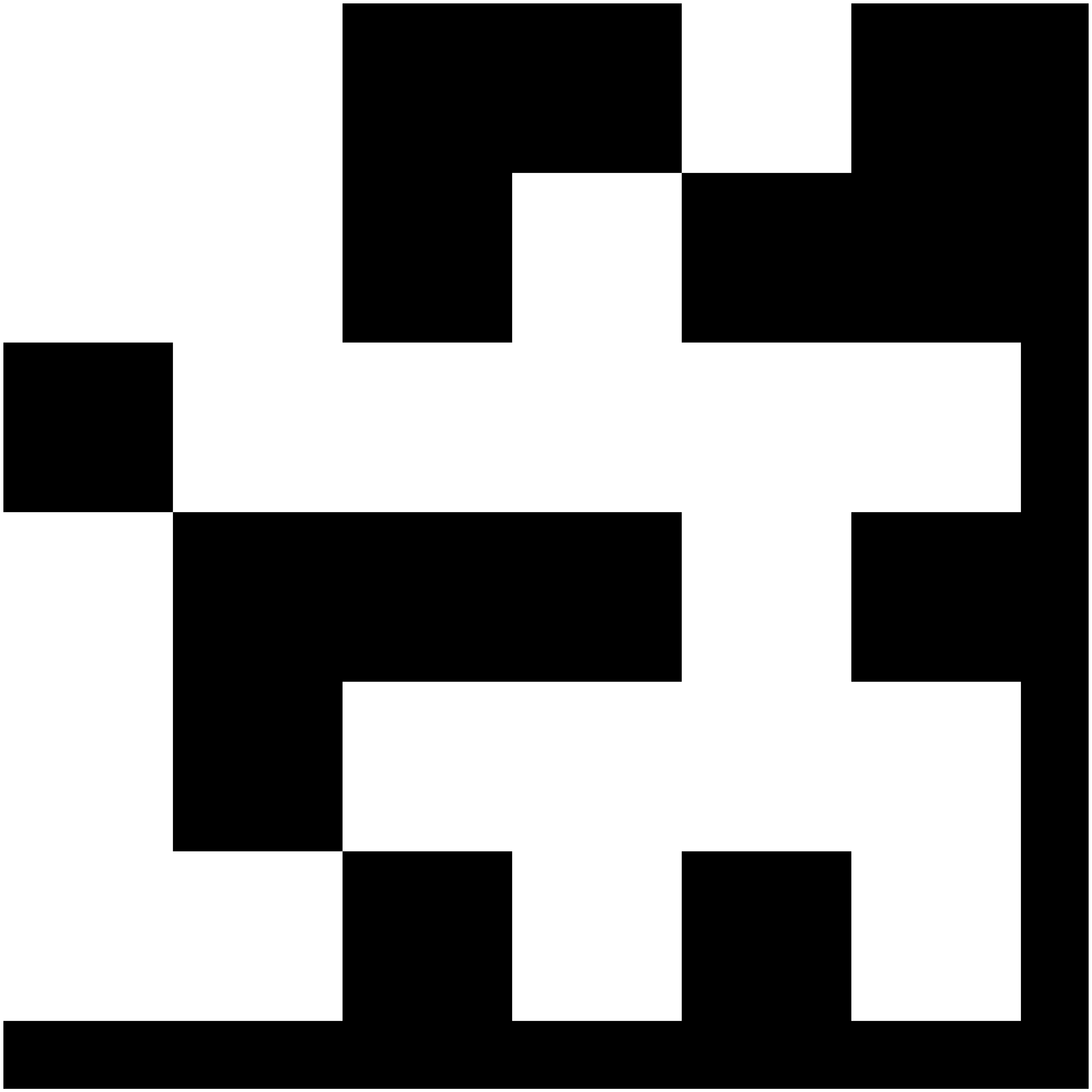}
    \caption{The first watermark for CNN model trained on the MNIST dataset. The dot code capacity of the image is 36 bits. The black border lines on the right and bottom are are due to the dimensions not being divisible by 6 without remains.}
    \label{fig:watermark_capacity_example}
\end{figure}

\vspace{0.1cm} \noindent 
\textbf{Capacity.} We showcase one of the images which are embedded into a model in \hyperref[eval:capacity]{\sect\ref{eval:capacity}}. The image shown in an additional figure in \hyperref[fig:watermark_capacity_example]{\fig\ref{fig:watermark_capacity_example}} represents a bit sequence.

\subsection{Additional Experiments}
\label{app:additionalexperiments}

\begin{figure}[tb]
\centering
\begin{subfigure}{0.48\linewidth}
    \includegraphics[width=\textwidth]{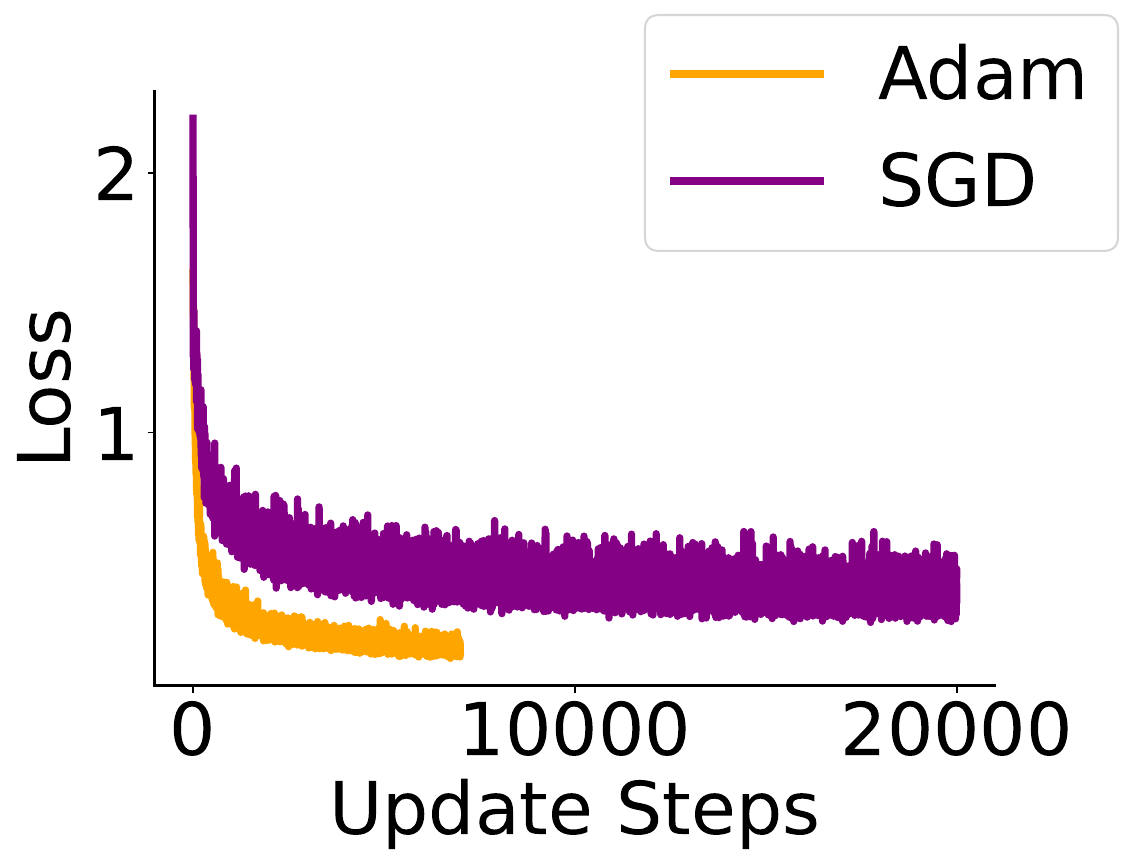}
    \vspace{-0.4cm}
\caption{}
\end{subfigure}
\hfill
\begin{subfigure}{0.48\linewidth}
    \includegraphics[width=\textwidth]{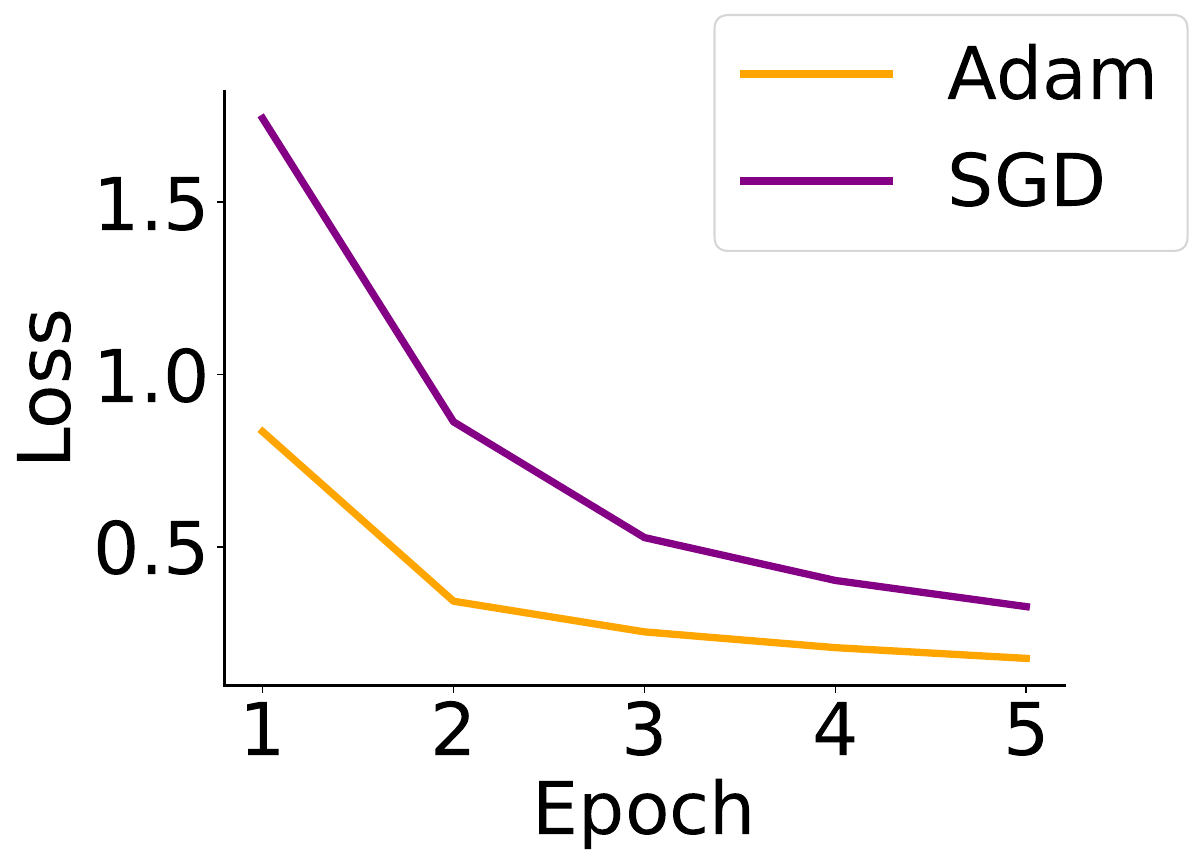}
    \vspace{-0.4cm}
\caption{}
\end{subfigure}

\caption{(a) SSIM loss applied in transposed training for Adam and SGD and (b) main task loss during watermark maintaining training for Adam and SGD.}
\label{fig:app:sgd}
\end{figure}

\vspace{0.1cm} \noindent 
\textbf{Different Optimizer.} To show the independence of \ourname from the optimizer, we conducted an experiment using SGD with a learning rate of 0.01. We terminated the watermark hardening after 20,000 update steps at an SSIM of 0.89 SSIM. We could achieve a main-task accuracy of 83.22\% after training while the SSIM remained high at 0.83. This result visualized in \hyperref[fig:app:sgd]{\fig\ref{fig:app:sgd}} shows, that \ourname functions for different optimizers with Adam outperforming SGD in both main-task and watermark embedding, which is an expected result. 

%\vspace{0.1cm} \noindent 
%\textbf{\cifar \& \resneteighteen.} For \cifar, we reached 0.96 SSIM after 4,000 update steps and 62.87\% accuracy with 0.96 SSIM after training. Fine-tuning yielded 67.64\% and 68.74\% accuracy with 0.69 and 0.55 SSIM for five and ten epochs respectively. The respective watermark after ten epochs can be seen in\hyperref[fig:applocks:exp10:a]{\fig\ref{fig:applocks:exp10:a}}.

\begin{figure}[tb]
\centering
\begin{subfigure}{0.13\linewidth}
    \includegraphics[width=\textwidth, cfbox=torstenbackground]{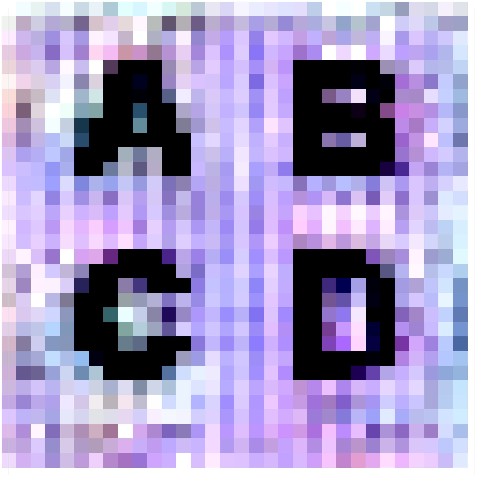} % 23_fc_mnist_3epochslowlr
    \vspace{-0.4cm}
\caption{}
    \label{fig:applocks:exp10:a_kannraus}
\end{subfigure}
\hfill
\begin{subfigure}{0.13\linewidth}
    \includegraphics[width=\textwidth, cfbox=torstenbackground]{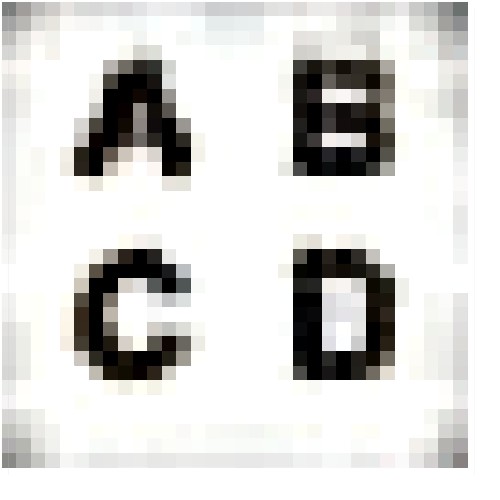} % 24_cnn_mnist_5epochs
    \vspace{-0.4cm}
\caption{}
    \label{fig:applocks:exp30:b}
\end{subfigure}
\hfill
\begin{subfigure}{0.13\linewidth}
    \includegraphics[width=\textwidth, cfbox=torstenbackground]{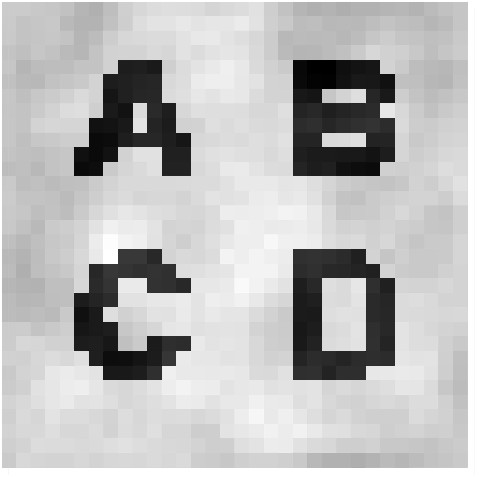} % 25_cnn_bn_mnist_5epochs
    \vspace{-0.4cm}
\caption{}
    \label{fig:applocks:exp14:a}
\end{subfigure}
\hfill
\begin{subfigure}{0.13\linewidth}
    \includegraphics[width=\textwidth, cfbox=torstenbackground]{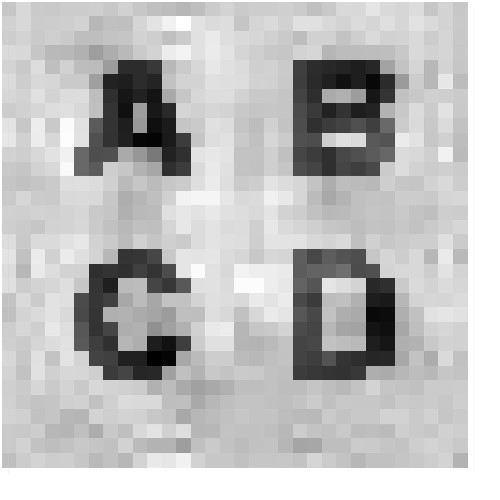} % 25_cnn_bn_mnist
    \vspace{-0.4cm}
\caption{}
    \label{fig:applocks:exp15:a}
\end{subfigure}
\hfill
\begin{subfigure}{0.13\linewidth}
    \includegraphics[width=\textwidth, cfbox=torstenbackground]{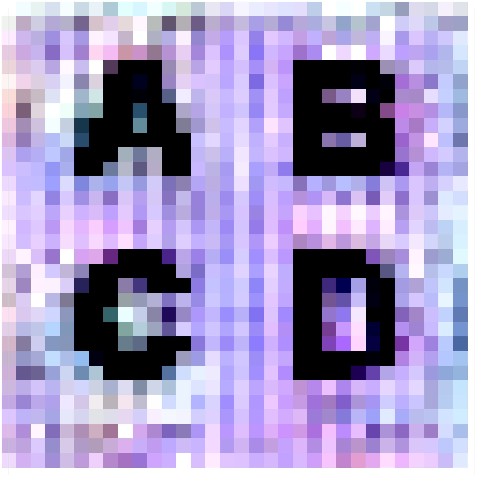} % 27_resnet34
    \vspace{-0.4cm}
\caption{}
    \label{fig:applocks:exp10cc:a}
\end{subfigure}
\caption{Extracted watermarks of (a) \cifar~\cite{cifar} on \resneteighteen\cite{resnet} after ten fine-tuning epochs with SSIM of 0.55}
\label{fig:applocks}
\end{figure}

\begin{figure}[tb]
\centering
\begin{subfigure}{\linewidth}
    \includegraphics[width=\textwidth]{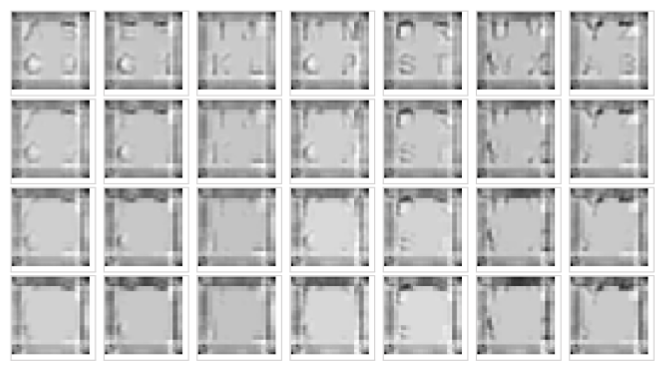} % 23_fc_mnist_3epochslowlr
\end{subfigure}
\caption{Extracted watermarks for an adaptive adversary that tries to erase an existing watermark key with an image that is extracted from an unwatermarked model. The four lines represent the extracted watermark after three to six adversarial update
steps with respective remaining accuracy values of 85.96\%, 83.07\%, 79.67\%, and 76.32\% originating from 89.10\% and SSIMs of 0.16, 0.08, 0.04, and 0.02.}
\label{fig:4}
\end{figure}

\vspace{0.1cm} \noindent 
\textbf{Adaptive Adversary.} An adversary could try to erase an embedded watermark. Thereby, an existing and already embedded key could be used in combination with the image that an unwatermarked model yields for that key for a transposed inference. We got similar results when using a completely black image to remove the watermark. As presented in \hyperref[eval:adaptive]{\sect\ref{eval:adaptive}} an adversary adapts \ournameGen training procedure and tries to erase the watermark. However, the entangled parameters force the adversary to sacrifice main task accuracy. After removing the watermark completely, the adversary sacrificed approximately 10\% accuracy in our setup compared to the original 89.10\%. In \hyperref[fig:4]{\fig\ref{fig:4}}, we show seven out of eleven watermark images after three to six adversarial update steps with respective remaining accuracy values of 85.96\%, 83.07\%, 79.67\%, and 76.32\%. For the first and second lines, the watermark can clearly be identified. In the third row, the watermark can be assumed, but it is already hard to discern, while in the fourth line, the watermark is mostly removed. Thereby, one should keep in mind, that an unwatermarked model yields an image similar to \hyperref[fig:1:exp2:a]{\fig\ref{fig:1:exp2:a}}.

\begin{figure}[tb]
\centering
\begin{subfigure}{\linewidth}
    \includegraphics[width=\textwidth]{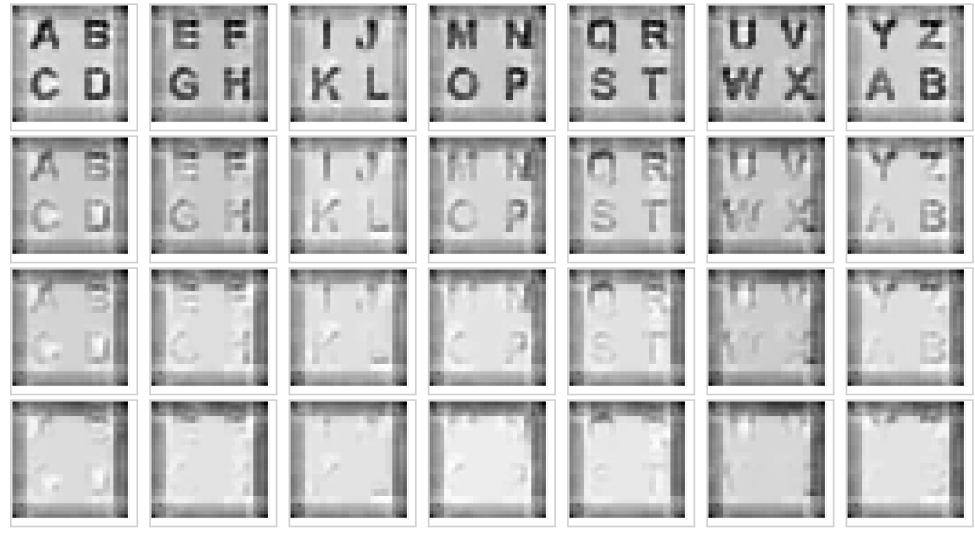} % 23_fc_mnist_3epochslowlr
\end{subfigure}
\caption{Extracted watermarks for an adaptive attack with a random key and an inferred image from an unwatermarked model as secret. The four lines show the extracted watermarks after the first four update steps. The adversary reduces the main task accuracy from 89.10\% to 88.61\%, 85.31\%, 81.00\%, and 74.81\% yielding SSIMs of 0.60, 0.26, 0.11, and 0.04.}
\label{fig:6}
\end{figure}

When using a random key and an inferred image from an unwatermarked model as a lock, we get similar results. Within the first four update steps, the adversary reduces the main task accuracy from 89.10\% to 88.61\%, 85.31\%, 81.00\%, and 74.81\%, essentially sacrificing more than 10\%. The yielded watermarks can be seen in the four rows of \hyperref[fig:6]{\fig\ref{fig:6}}, showing that the watermark is still clearly visible in the third row. Even in row four, one can see the watermark slightly. The results for a random key combined with a black image or a random image show the same effect as reported in this experiment.

\balance

\end{document}